%% file: main.tex
\pdfoutput=1
\documentclass[10pt,twocolumn,letterpaper]{article}
\usepackage[pagenumbers]{cvpr}
\usepackage{multirow}
\usepackage{float}
\usepackage{adjustbox}
\usepackage{pgfplots}
\pgfplotsset{compat=1.17}
\usetikzlibrary{arrows.meta,positioning,patterns}
\definecolor{cvprblue}{rgb}{0.21,0.49,0.74}
\usepackage[breaklinks,colorlinks,allcolors=cvprblue]{hyperref}
\hypersetup{pdftitle={Shortcut or Overlap? Separating Demographic Reliance from Genuine Biology in Medical Vision Foundation Models},pdfauthor={Ritvij Sharma}}

\makeatletter
\def\@maketitle{%
  \newpage\null
  \vskip 0.05in
  \hrule height 4pt
  \vskip 0.25in
  \begin{center}
    {\LARGE\bf \@title\par}
  \end{center}
  \vskip 0.22in
  \hrule height 1pt
  \vskip 0.32in
  \begin{center}
    \begin{tabular}[t]{c}\@author\end{tabular}\par
  \end{center}
  \vskip 0.3in
}
\makeatother

\title{Shortcut or Overlap? Separating Demographic Reliance\\from Genuine Biology in Medical Vision Foundation Models}

\author{\textbf{Ritvij Sharma}\\
Starspark.ai\\
\texttt{i-ritvij@starspark.ai}
}

\begin{document}
\maketitle
\begin{abstract}
If age and sex are removed from a medical foundation model's features, its disease predictions usually get worse. This drop has two very different explanations. The model may use age or sex as a shortcut, which is a fairness problem. Alternatively, the disease may change the image in the same way that aging does, as glaucoma and aging both change the optic disc, which is sound biology. Standard erasure analyses cannot tell these apart. We propose a test that can. It compares diseased and healthy patients of the same age and sex, so that demographics no longer predict the diagnosis, and asks whether removing age and sex from the features still hurts. A control that removes randomly reassigned demographics calibrates the answer. In experiments with known answers, the test ignores age confounding that we add to a dataset and detects disease signal that we plant along the age direction. We apply it to six foundation models on retinal photographs and chest radiographs. After matching on age and sex, no measured cost exceeds the test's resolution of 0.005 AUC, so each is best explained as a shortcut. On PAPILA, removing age and sex costs up to 0.09 AUC, and the cost is no longer detectable after matching 79 pairs on age. On ODIR-5K and for cardiomegaly, matching on age leaves a residual that disappears once sex is matched as well. Fine-tuning can deepen a shortcut: on PAPILA it leaves AUC unchanged yet nearly doubles how strongly the glaucoma score tracks age in healthy eyes.
\end{abstract}
\begin{figure}[t]
\centering
\begin{tikzpicture}
\begin{axis}[width=\columnwidth, height=5.0cm, ybar=1pt, bar width=7pt,
  symbolic x coords={P,O,E,C}, xtick=data, enlarge x limits=0.15,
  xticklabels={{PAPILA\\glaucoma\\RETFound},{ODIR-5K\\glaucoma\\RETFound},{NIH CXR\\effusion\\DINOv2-L},{NIH CXR\\cardiomeg.\\DINOv2-L}},
  xticklabel style={align=center, font=\scriptsize}, ymin=-0.068, ymax=0.032,
  ylabel={AUC change}, ylabel style={font=\footnotesize}, yticklabel style={font=\scriptsize, /pgf/number format/fixed, /pgf/number format/precision=2},
  scaled y ticks=false, ymajorgrids, grid style={gray!25},
  legend style={font=\scriptsize, at={(0.5,1.02)}, anchor=south, legend columns=1, draw=none, cells={anchor=west}},
  axis x line*=bottom, axis y line*=left, extra y ticks={0}, extra y tick style={grid style={black!60}}]
\addplot[fill=gray!45, draw=gray!70] coordinates {(P,-0.057) (O,-0.022) (E,-0.016) (C,-0.023)};
\addplot[fill=orange!75!red, draw=orange!60!black, postaction={pattern=north east lines, pattern color=orange!40!black}] coordinates {(P,0.017) (O,-0.008) (E,0.000) (C,-0.008)};
\addplot[fill=blue!55, draw=blue!60!black, postaction={pattern=crosshatch dots, pattern color=blue!30!black}] coordinates {(P,0) (O,-0.0003) (E,0.0003) (C,-0.0002)};
\legend{Erasure cost on all patients, Excess after age matching, Excess after age and sex matching}
\node[font=\tiny, text=gray!70!black, xshift=8pt, rotate=90, anchor=west] at (axis cs:P,0.001) {n/a};
\end{axis}
\end{tikzpicture}
\caption{\textbf{Large erasure costs, all explained by demographics.} Erasing sex and age from a foundation model's features lowers disease AUC in every setting (gray), which by itself cannot distinguish a demographic shortcut from shared biology. After matching on age (orange, hatched), PAPILA and effusion show no excess cost over a shuffled-demographic null. ODIR-5K and cardiomegaly keep a residual, which disappears once sex is matched as well (blue, dotted). n/a: not run on PAPILA (79 pairs).}
\label{fig:teaser}
\end{figure}
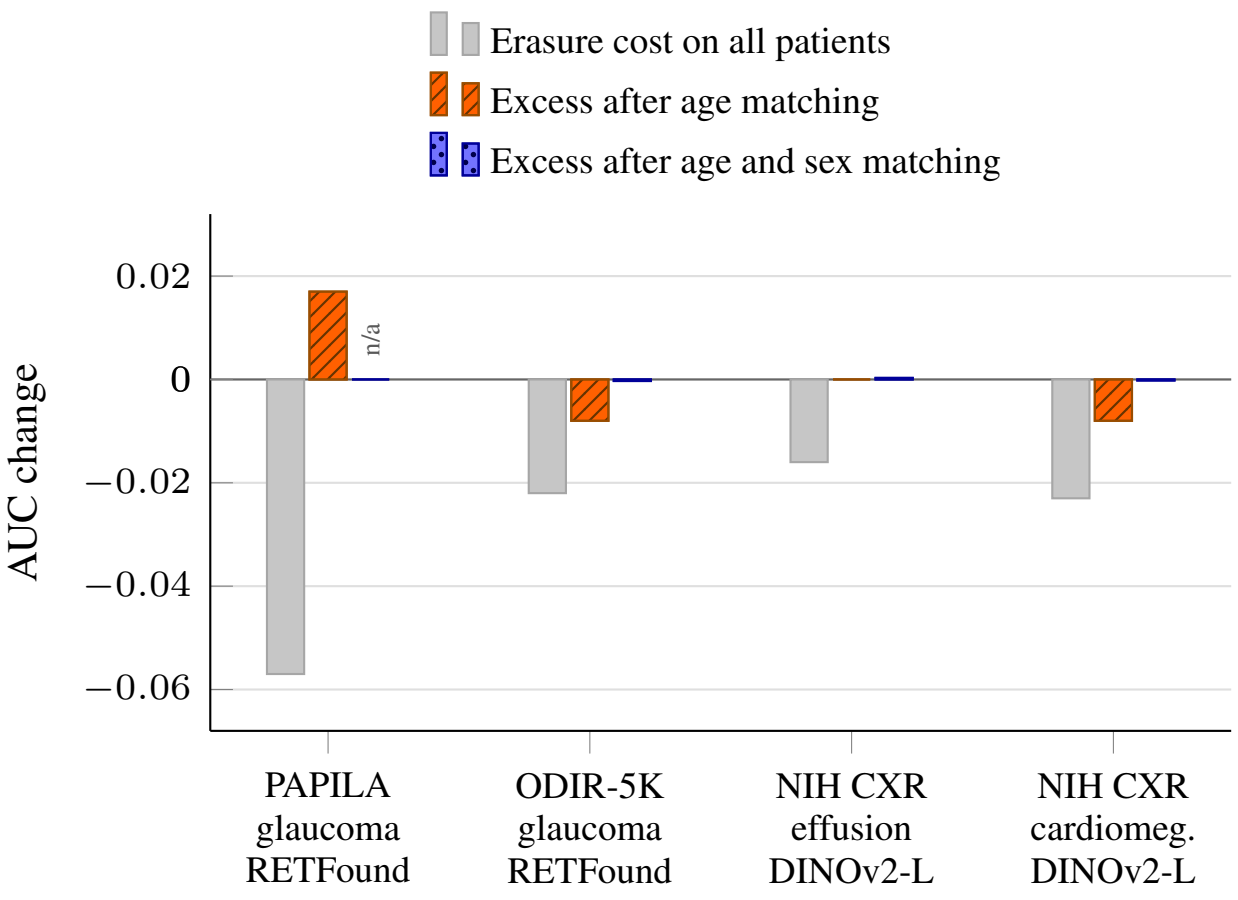

\section{Introduction}
\label{sec:intro}

Foundation models for retinal photographs and chest radiographs detect disease from frozen features with little task-specific training~\cite{zhou2023retfound,perezgarcia2025raddino}. They also encode who the patient is. Networks predict age and sex from fundus photographs~\cite{munk2021patient} and self-reported race from radiographs~\cite{gichoya2022race}, and a chest X-ray foundation model was shown to encode sex and race in ways that produce uneven subgroup performance~\cite{glocker2023cxrfm}. Before such a model helps decide who is referred for care, an auditor needs to know whether its disease predictions use that information and, if they do, why.

The standard tool is concept erasure~\cite{ravfogel2020inlp,belrose2023leace}. A foundation model turns each image into a feature vector, and age and sex each correspond to a direction in that vector space. Erasure removes those directions, retrains a simple disease classifier on what remains, and measures how much accuracy is lost. A loss is usually read as evidence of a shortcut~\cite{geirhos2020shortcut}. In medicine this reading can be wrong, because demographics can be causally tied to disease~\cite{brown2023short}. Glaucoma becomes far more common with age~\cite{tham2014glaucoma}, and aging and glaucoma both change the optic disc. Suppose removing age lowers glaucoma accuracy. The model may be treating older-looking retinas as glaucomatous, which we call a \emph{shortcut}; it is a fairness problem, because a healthy older patient would be flagged. Or the visual signs of glaucoma may partly resemble the visual signs of age, so that removing age also removes some disease signal. We call this an \emph{overlap}; it reflects real biology. Both explanations produce the same drop (\cref{fig:teaser}, gray).

The two explanations make different predictions if age and sex stop predicting the diagnosis. A shortcut relies on that association, so it should vanish. An overlap lives in the image features themselves, so it should persist. We therefore pair each diseased patient with a healthy patient of the same sex and similar age, retrain the classifier on these pairs, and erase again. Two controls make the comparison fair. Removing \emph{any} two directions costs a little accuracy, so we also remove directions built from age and sex randomly reassigned between patients, and report only the loss beyond this baseline, which we call the \emph{excess cost}. A single pairing can be lucky, so we repeat the test over 20 random pairings. We check the test on cases with known answers. It gives the same verdict after we add age confounding to a dataset, and it detects an overlap that we plant in the features.

We audit three retinal foundation models (RETFound~\cite{zhou2023retfound}, Dual-IFM~\cite{dualifm2026}, and the general-purpose DINOv2~\cite{oquab2024dinov2}) on two glaucoma datasets with different demographic structure, and three models on chest radiographs (RAD-DINO~\cite{perezgarcia2025raddino} and two DINOv2 variants). Matching on age alone removes the cost on PAPILA and for effusion. For glaucoma on ODIR-5K and for cardiomegaly, it leaves a consistent residual that an age-only analysis would call an overlap. Matching on sex as well removes that residual in every model. Our contributions are:
\begin{itemize}[leftmargin=*,itemsep=1pt,topsep=2pt]
\item A matched erasure test with a shuffled-demographic null and a multi-matching excess statistic that separates demographic shortcuts from overlap in frozen foundation models, a score-only variant for auditors without feature access, and known-answer experiments that measure its behavior.
\item An audit of six foundation models across three datasets and two imaging domains under shared folds, showing that no erasure cost exceeds the test's resolution after age and sex matching and that matching on age alone would misattribute sex shortcuts to biology.
\item Evidence that adaptation and erasure leave shortcuts in place. On a small, age-confounded dataset, fine-tuning nearly doubles age dependence across three seeds at unchanged AUC, and linear erasure that passes linear checks leaves age readable to an MLP.
\end{itemize}
\section{Related Work}
\label{sec:related}

\noindent\textbf{Medical foundation models and demographics.}
RETFound~\cite{zhou2023retfound} was pretrained by self-supervision on 1.6 million retinal images. RETFound-Green~\cite{engelmann2025retfoundgreen} trains a comparable retinal model on 75,000 public images with 400 times less compute, and Dual-IFM~\cite{dualifm2026} is a retinal foundation model designed for interpretability; we use its released t-SimCNE-2D checkpoint, which has a ResNet-50 backbone. RAD-DINO~\cite{perezgarcia2025raddino} adapts DINOv2~\cite{oquab2024dinov2} to chest radiographs using images alone. Prior audits establish that these images carry demographic signal. Deep networks predict age and sex from fundus photographs~\cite{munk2021patient}, retinal-age errors of fine-tuned RETFound differ across sex and ethnicity groups~\cite{nielsen2025retinalage}, and chest X-ray models encode sex and race~\cite{gichoya2022race,glocker2023cxrfm}. These studies measure what is encoded or how performance varies across groups. They do not ask whether a disease prediction uses demographics, or whether that use reflects a shortcut or shared biology.

\noindent\textbf{Shortcut learning.}
Geirhos \etal~\cite{geirhos2020shortcut} define shortcuts as decision rules that succeed on standard benchmarks and fail under distribution shift. The closest work to ours is ShorT~\cite{brown2023short}, which uses multitask training to vary how strongly a model encodes a sensitive attribute and tracks the resulting change in fairness, explicitly because the attribute may be causally linked to disease. ShorT retrains task models. We audit frozen foundation models, control age by matching instead of retraining, and compare models and datasets under identical folds.

\noindent\textbf{Probing and concept erasure.}
Linear probes measure what a layer encodes~\cite{alain2016probes}. Iterative nullspace projection~\cite{ravfogel2020inlp} and LEACE~\cite{belrose2023leace} prevent any linear classifier from recovering a concept. In language models, kernelized erasure~\cite{ravfogel2022kernel} showed that protection against one non-linear adversary does not transfer to others. We use erasure as an intervention, add a matched design and a null to interpret its cost, and test whether the linear guarantee leaves age readable in medical image features. Top-$k$ sparse autoencoders (SAEs)~\cite{gao2024sae} recover biological concepts in a pathology foundation model~\cite{le2024pathsae} but have underperformed linear probes on safety tasks~\cite{smith2025saeneg}; we use them only as a diagnostic.
\begin{figure*}[t]
\centering
\begin{tikzpicture}[font=\footnotesize, node distance=4.5mm,
  box/.style={draw, rounded corners=2pt, align=left, inner sep=3.5pt, fill=#1},
  box/.default=gray!6, arr/.style={-{Stealth[length=2mm]}, thick}]
\node[box=gray!10, text width=3.4cm] (in) {\textbf{Frozen features} $x$\\[1pt]
 Retina: RETFound, DINOv2, Dual-IFM; ODIR-5K, PAPILA\\[1pt]
 CXR: RAD-DINO, DINOv2-B/L; cardiomegaly, effusion\\[1pt]
 \textit{shared patient-grouped folds}};
\node[box=blue!6, text width=4.5cm, right=of in, yshift=8mm] (ins) {\textbf{Inside audit} (\cref{sec:erasure})\\[1pt]
 probe $f(x)\to y$\\
 erase sex and age: $\tilde x = Px$ \hfill(\cref{eq:proj})\\
 cost $\Delta$ vs.\ shuffled null \hfill(\cref{eq:delta,eq:pval})};
\node[box=orange!8, text width=4.5cm, right=of in, yshift=-13mm] (out) {\textbf{Outside-in audit} (\cref{sec:outside})\\[1pt]
 scores only: $\rho$(score, age $\mid y{=}0$)\\
 AUC drop on matched set \hfill(\cref{eq:outside})};
\node[box=green!7, text width=4.9cm, right=of ins, yshift=-8mm] (mt) {\textbf{Matched test} (\cref{sec:matched})\\[1pt]
 pair $y{=}1$ with $y{=}0$: same sex, age $\pm2$\,y \hfill(\cref{eq:match})\\
 retrain and re-erase on $\mathcal M\to\Delta_{\mathcal M}$\\
 subtract null on same $\mathcal M\to E$ \hfill(\cref{eq:excess})\\
 repeat over 20 matchings $\to\bar E$, $n_-$};
\node[box=orange!20, text width=2.2cm, right=of mt, yshift=7mm, align=center] (s) {\textbf{Shortcut}\\ $\bar E\ge0$\\ cost vanishes};
\node[box=blue!14, text width=2.2cm, right=of mt, yshift=-7mm, align=center] (o) {\textbf{Overlap}\\ $n_-\ge18$/20\\ cost persists};
\draw[arr] (in.east) -- ++(2mm,0) |- (ins.west);
\draw[arr] (in.east) -- ++(2mm,0) |- (out.west);
\draw[arr] (ins.east) -- (mt.west |- ins.east);
\draw[arr] (mt.east) -- (s.west);
\draw[arr] (mt.east) -- (o.west);
\draw[arr, dashed, gray] (out.east) -- (mt.west |- out.east);
\end{tikzpicture}
\caption{\textbf{Overview of the audit.} Frozen features feed an inside audit (probe, erasure, null) and a score-only outside-in audit. When erasure costs AUC, the matched test retrains and re-erases on patients matched for sex and age and compares the result with a shuffled-demographic null over 20 matchings, which separates a shortcut from an overlap. The pipeline is rerun on features from fine-tuned models (\cref{sec:ftmethod}).}
\label{fig:pipeline}
\end{figure*}
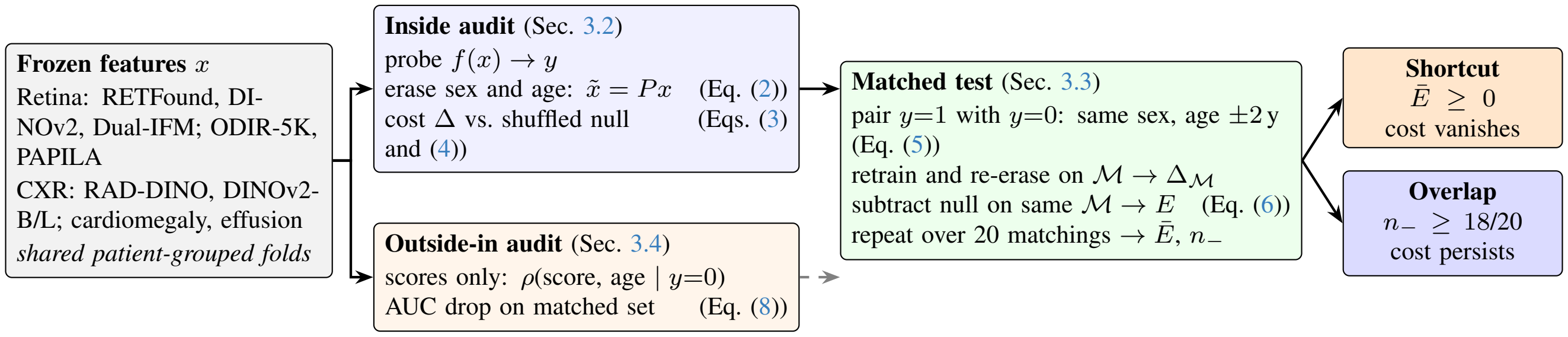

\section{Method}
\label{sec:method}

\subsection{Problem Formulation}
\label{sec:formulation}
Let $x_i\in\mathbb{R}^d$ be the frozen feature of image $i$ (patch tokens averaged at a fixed layer), $y_i\in\{0,1\}$ a disease label, $a_i$ age, $s_i$ sex, and $g_i$ the patient. A probe $f$, the simple disease classifier of \cref{sec:intro}, is an $\ell_2$-regularized logistic regression ($C{=}1$, standardized inputs) trained and evaluated with patient-grouped 5-fold cross-validation, so that no patient appears in both training and test data. The folds of every main analysis are fixed once per dataset and shared by every model and method; the robustness checks of the RETFound and Dual-IFM ODIR-5K audits build their own patient-grouped split (\cref{sec:supp_impl}).

A model \emph{relies} on demographics for $y$ if removing the linearly decodable demographic information from $x$ lowers the out-of-fold AUC of $f$. Reliance can arise in two ways. A \emph{shortcut} arises when the probe exploits an association between demographics and the label in the sample, $p(y\mid a,s)\neq p(y)$, so that the demographic directions help only because age or sex predicts disease in these data. An \emph{overlap} arises when the disease direction in feature space shares variance with the demographic directions, so that erasure removes disease signal even where age and sex carry no information about the label. We seek to distinguish the two using frozen features alone (\cref{fig:pipeline}). We measure accuracy by the area under the ROC curve (AUC), and report changes in AUC.

\subsection{Demographic Erasure and Shuffled Null}
\label{sec:erasure}
On each training fold $\mathcal T$ we compute the mean-difference directions
\begin{equation}
\begin{aligned}
d_s&=\mu_{\mathcal T}(x\mid s{=}1)-\mu_{\mathcal T}(x\mid s{=}0),\\
d_a&=\mu_{\mathcal T}(x\mid a{\ge}60)-\mu_{\mathcal T}(x\mid a{<}60),
\end{aligned}
\label{eq:dirs}
\end{equation}
orthonormalize them into $D=[\hat d_s,\hat d_a]^\top$, and project them out of both training and test features:
\begin{equation}
\tilde x = P x,\qquad P = I - D^\top D .
\label{eq:proj}
\end{equation}
Retraining the probe on $\tilde x$ gives the erasure cost
\begin{equation}
\Delta=\mathrm{AUC}\big(\tilde f(\tilde x),y\big)-\mathrm{AUC}\big(f(x),y\big),
\label{eq:delta}
\end{equation}
which is negative when erasure hurts. LEACE~\cite{belrose2023leace} on sex and standardized continuous age serves as a second erasure that is provably complete for linear classifiers. Because removing any two directions can cost accuracy, we compare $\Delta$ with a null. We permute $(s,a)$ across patients with permutations $\pi_k$, $k=1,\dots,N$ ($N{=}100$), rebuild the directions identically, and compute $\Delta^{(\pi_k)}$:
\begin{equation}
p=\frac{1+\sum_{k}\mathbb 1\big[\Delta^{(\pi_k)}\le\Delta\big]}{1+N}.
\label{eq:pval}
\end{equation}

\subsection{The Matched Shortcut Test}
\label{sec:matched}
A shortcut requires $p(y\mid a,s)\neq p(y)$ and an overlap does not, so we remove that association by design. Visiting positives in random order, we greedily pair each one, without replacement, with a negative of the same sex whose age differs by at most $c{=}2$ years:
\begin{equation}
\mathcal M=\{(i,j): y_i{=}1,\ y_j{=}0,\ s_i{=}s_j,\ |a_i-a_j|\le c\}.
\label{eq:match}
\end{equation}
Within $\mathcal M$, age and sex carry almost no information about $y$. We also run an age-only variant that drops the condition $s_i{=}s_j$; a residual that survives age-only matching but disappears under joint matching is a sex shortcut. We retrain and re-erase on $\mathcal M$ alone, with new grouped folds, to obtain the matched cost $\Delta_{\mathcal M}$. A shortcut predicts $\Delta_{\mathcal M}\approx0$, and an overlap predicts $\Delta_{\mathcal M}<0$.

\noindent\textbf{Excess over the null.}
$\Delta_{\mathcal M}$ still contains the generic cost of removing two directions, which is large in dense early-layer features. We subtract the mean shuffled-demographic cost computed on the same matched set:
\begin{equation}
E=\Delta_{\mathcal M}-\frac1K\sum_{k=1}^{K}\Delta_{\mathcal M}^{(\pi_k)},\qquad K=20 .
\label{eq:excess}
\end{equation}

\noindent\textbf{Multiple matchings.}
A single greedy matching can be favorable by chance, so we repeat the matching with $R{=}20$ seeds and report $\bar E=\frac1R\sum_r E_r$ and $n_-=\#\{r:E_r<0\}$. The matchings reuse patients and are therefore dependent, so $n_-$ measures stability, not significance. We label a residual an \emph{overlap} when, after joint matching, $n_-\ge18$ and $|\bar E|\ge0.001$ AUC, and a \emph{shortcut} when $\Delta$ is significant but $\bar E\ge0$ or $n_-\le10$; $10<n_-<18$ is inconclusive, and no excess within the 0.005 AUC resolution is read as overlap. \Cref{sec:known} measures how small an excess the test can resolve.

\noindent\textbf{Where the cost lies.}
As a complementary check we compute the erasure cost within age bands $b\in\{{<}60,\ 60\text{--}69,\ {\ge}70\}$, weighted by the share of positives $w_b$ in each band:
\begin{equation}
\Delta=\underbrace{\textstyle\sum_b w_b\,\Delta_b}_{\Delta_{\text{within}}}+\Delta_{\text{between}} .
\label{eq:bands}
\end{equation}
A shortcut concentrates the cost between bands, and an overlap concentrates it within them.

\subsection{Outside-in Audit}
\label{sec:outside}
An auditor may have access to scores $\hat y$ only. Two score-only statistics mirror the test:
\begin{equation}
\begin{aligned}
\rho&=\mathrm{Spearman}\big(\hat y_i,a_i \,\big|\, y_i{=}0\big),\\
D&=\mathrm{AUC}_{\text{all}}(\hat y)-\mathrm{AUC}_{\mathcal M}(\hat y).
\end{aligned}
\label{eq:outside}
\end{equation}
Among disease-free patients, a real age--disease link cannot account for a correlation between score and age. $D$ measures how much of the AUC depends on age differences between patients.

\subsection{Fine-tuning Protocol}
\label{sec:ftmethod}
To test whether shortcuts survive adaptation, we partially fine-tune RETFound (the last 4 of 24 blocks, the final norm, and a linear head) inside the same outer folds, with a fixed number of epochs and no early stopping on test data. For each fold we re-extract features for every image with that fold's model and rerun \cref{sec:erasure,sec:matched,sec:outside}, training every probe only on that fold's training images.
\section{Experiments}
\label{sec:exp}

\subsection{Experimental Settings}
\label{sec:setup}

\noindent\textbf{Datasets.}
\emph{ODIR-5K}~\cite{odir2019} contains 6,713 fundus images (both eyes of 3,358 patients) with age, sex, and per-eye diagnostic keywords. We derive labels for glaucoma (316 eyes, 42 of them recorded as suspected), cataract, and five other conditions with explicit keyword rules. Negatives are all eyes without the label, and age is only weakly associated with glaucoma. \emph{PAPILA}~\cite{kovalyk2022papila} provides clinical diagnoses for 420 eyes of 210 patients after excluding suspects (87 glaucoma, 333 healthy). Its healthy controls are younger than its patients, so age is confounded with diagnosis by construction. From \emph{NIH ChestX-ray14}~\cite{wang2017chestxray8} we keep one frontal image per patient aged 18 to 95. Cardiomegaly (1,496 positives) and pleural effusion (3,033 positives) are each contrasted with 8,000 ``No Finding'' images, and positives are on average 5 years older than negatives (51 \vs 46).

\noindent\textbf{Models.}
For retina we use RETFound~\cite{zhou2023retfound} (ViT-L/16, retinal), DINOv2~\cite{oquab2024dinov2} (ViT-L/14, natural images), and Dual-IFM~\cite{dualifm2026} (ResNet-50, retinal). For chest radiographs we use RAD-DINO~\cite{perezgarcia2025raddino} (ViT-B/14 initialized from DINOv2-B and trained on chest radiographs), DINOv2-B (the same architecture trained on natural images), and DINOv2-L. The main layer sits at two-thirds of the depth (block 16 of 24, block 8 of 12, or the third residual stage), and one-third and final depth serve as robustness checks. All models are frozen except in the fine-tuning experiment.

\noindent\textbf{Metrics and implementation.}
We report AUC and the erasure costs $\Delta$, $\Delta_{\mathcal M}$, and $\bar E$ in AUC units, with 95\% percentile intervals from 1,000 patient-level bootstrap resamples, paired for differences. $p$-values are uncorrected and descriptive. Robustness checks use 100 nulls for linear probes and 30 for an MLP probe with one hidden layer of 256 units. Retinal images enter each model at its pretraining resolution (224 pixels; 256 for Dual-IFM), and chest radiographs pass through each model's own preprocessing (518 pixels for RAD-DINO). Fine-tuning uses AdamW with learning rates of $10^{-5}$ for the backbone and $10^{-3}$ for the head, a one-cycle schedule, batch size 16, and 15 (PAPILA) or 8 (ODIR-5K) epochs. \Cref{sec:supp_impl} lists every remaining hyperparameter.
\subsection{Main Results: Shortcut or Overlap?}
\label{sec:main}

\begin{figure*}[t]
\centering
\begin{tikzpicture}
\pgfplotsset{
  every axis/.style={width=0.47\textwidth, height=7.6cm, xmajorgrids, grid style={gray!25},
    tick label style={font=\footnotesize}, scaled x ticks=false, xticklabel style={/pgf/number format/fixed, /pgf/number format/precision=3}, label style={font=\footnotesize}, title style={font=\footnotesize},
    ytick=data, y dir=reverse, enlarge y limits=0.06},
}
\begin{axis}[name=A, symbolic y coords={OR,OD,OI,PR,PD,PI,CR,CB,CL,ER,EB,EL},
  yticklabels={ODIR glauc.\ RETFound, ODIR glauc.\ DINOv2, ODIR glauc.\ Dual-IFM,
               PAPILA glauc.\ RETFound, PAPILA glauc.\ DINOv2, PAPILA glauc.\ Dual-IFM,
               CXR cardiomeg.\ RAD-DINO, CXR cardiomeg.\ DINOv2-B, CXR cardiomeg.\ DINOv2-L,
               CXR effusion RAD-DINO, CXR effusion DINOv2-B, CXR effusion DINOv2-L},
  xmin=-0.1, xmax=0.02, xlabel={Erasure cost $\Delta$ (AUC)},
  title={(a) All patients vs.\ age-matched (mean of 20 matchings)},
  legend style={font=\footnotesize, at={(0.02,0.98)}, anchor=north west, draw=none, fill=white, fill opacity=0.8, text opacity=1}]
\draw[gray, dashed] (axis cs:0,OR) -- (axis cs:0,EL);
\addplot[only marks, mark=o, mark size=2.2pt, color=blue!70!black] coordinates {
 (-0.022,OR) (-0.008,OD) (-0.021,OI) (-0.057,PR) (-0.090,PD) (-0.032,PI)
 (-0.0161,CR) (-0.0259,CB) (-0.0226,CL) (-0.0133,ER) (-0.0218,EB) (-0.0164,EL)};
\addplot[only marks, mark=*, mark size=2.2pt, color=orange!90!black] coordinates {
 (-0.012,OR) (-0.001,OD) (-0.005,OI) (0.0099,PR) (-0.0033,PD) (0.0119,PI)
 (-0.0034,CR) (-0.0076,CB) (-0.0096,CL) (-0.0006,ER) (-0.0011,EB) (-0.0008,EL)};
\legend{all patients, age-matched}
\end{axis}
\begin{axis}[at={(A.east)}, anchor=west, xshift=0.3cm, symbolic y coords={OR,OD,OI,PR,PD,PI,CR,CB,CL,ER,EB,EL},
  yticklabels={}, xmin=-0.011, xmax=0.021, xtick={-0.005,0,0.005,0.01,0.015,0.02}, xlabel={Excess cost $\bar E$ over shuffled null (AUC)},
  title={(b) Excess beyond chance after matching},
  legend style={font=\footnotesize, at={(0.985,0.015)}, anchor=south east, legend columns=1, draw=gray!50, fill=white, cells={anchor=west}}]
\fill[gray!14] ([yshift=7pt]axis cs:-0.005,OR) rectangle ([yshift=-7pt]axis cs:0.005,EL);
\draw[gray, dashed] ([yshift=7pt]axis cs:0,OR) -- ([yshift=-7pt]axis cs:0,EL);
\draw[gray!70, thick, -{Stealth[length=2mm]}, shorten >=3pt, shorten <=3pt] (axis cs:-0.008,OR) -- (axis cs:-0.0003,OR);
\draw[gray!70, thick, -{Stealth[length=2mm]}, shorten >=3pt, shorten <=3pt] (axis cs:0.001,OD) -- (axis cs:-0.0009,OD);
\draw[gray!70, thick, -{Stealth[length=2mm]}, shorten >=3pt, shorten <=3pt] (axis cs:-0.003,OI) -- (axis cs:0.0005,OI);
\draw[gray!70, thick, -{Stealth[length=2mm]}, shorten >=3pt, shorten <=3pt] (axis cs:-0.0017,CR) -- (axis cs:0.0005,CR);
\draw[gray!70, thick, -{Stealth[length=2mm]}, shorten >=3pt, shorten <=3pt] (axis cs:-0.0059,CB) -- (axis cs:-0,CB);
\draw[gray!70, thick, -{Stealth[length=2mm]}, shorten >=3pt, shorten <=3pt] (axis cs:-0.008,CL) -- (axis cs:-0.0002,CL);
\addplot[only marks, mark=diamond*, mark size=2.8pt, color=red!70!black] coordinates {
 (-0.008,OR) (0.001,OD) (-0.003,OI) (0.0167,PR) (0.0033,PD) (0.0102,PI)
 (-0.0017,CR) (-0.0059,CB) (-0.0080,CL) (0.0003,ER) (-0.0004,EB) (0.0000,EL)};
\addplot[only marks, mark=square*, mark size=2.1pt, color=blue!70!black] coordinates {
 (-0.0003,OR) (-0.0009,OD) (0.0005,OI) (0.0005,CR) (-0.0000,CB) (-0.0002,CL) (0.0004,ER) (-0.0004,EB) (0.0003,EL)};
\addlegendimage{area legend, fill=gray!14, draw=none}
\legend{after age matching, after age and sex matching, test resolution ($\pm0.005$)}
\end{axis}
\end{tikzpicture}
\caption{\textbf{Separating shortcut from overlap.} (a) Erasing sex and age lowers AUC in every setting (open circles); after age matching (filled, mean of 20 matchings), the PAPILA and effusion costs disappear. (b) Excess cost beyond the shuffled null: red after age matching, blue after age and sex matching (not run on PAPILA), arrows showing the change. Points inside the gray band, the test's resolution (\cref{sec:known}), show no detectable overlap. Matching counts are in \cref{tab:main}.}
\label{fig:main}
\end{figure*}
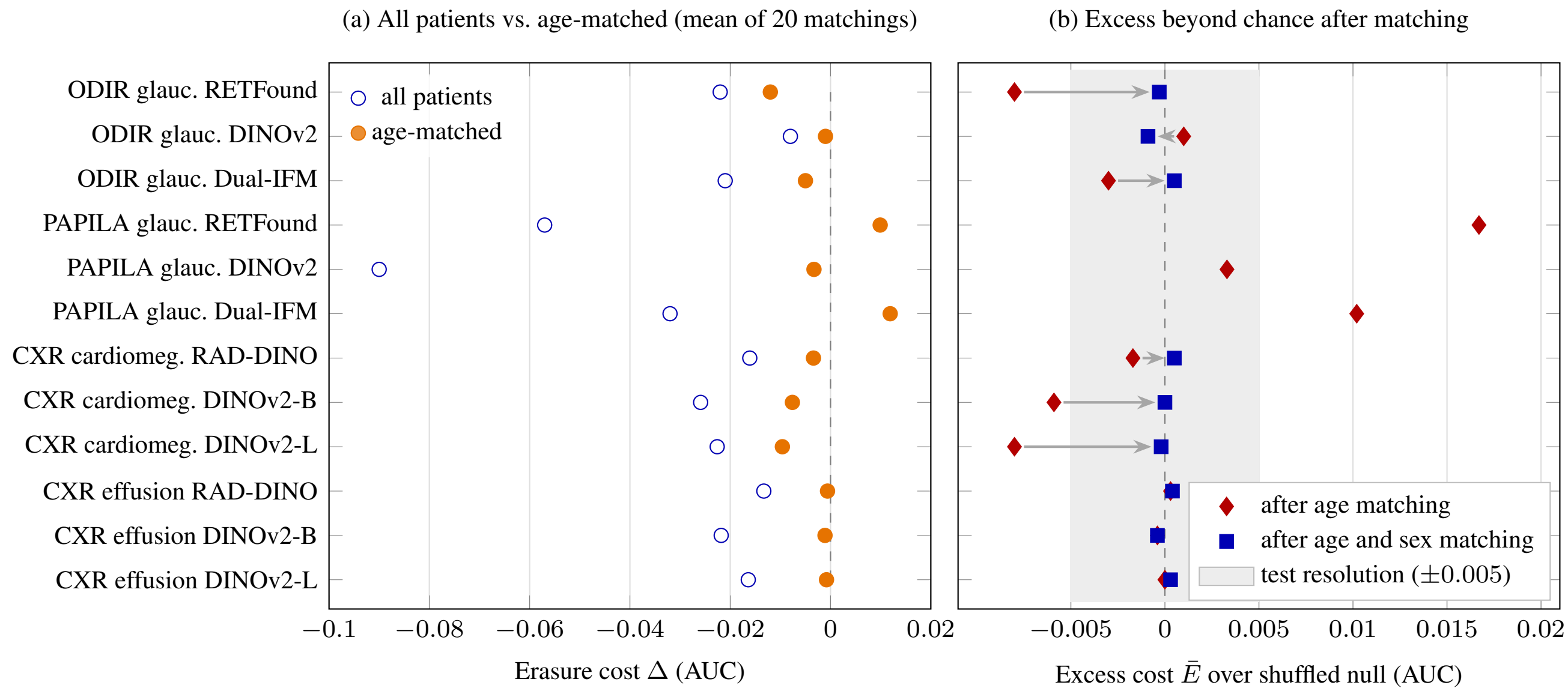

\begin{table}[t]
\centering\footnotesize\setlength{\tabcolsep}{2.2pt}
\adjustbox{max width=\columnwidth}{%
\begin{tabular}{@{}llrrrrl@{}}
\toprule
 & & & \multicolumn{2}{c}{$\bar E$ ($E{<}0$ of 20)} & & \\
\cmidrule(lr){4-5}
Setting & Model & $\Delta$ all & age & age + sex & $\Delta_{\mathcal M}$ & Shortcut via \\
\midrule
\multirow{3}{*}{\shortstack[l]{ODIR\\glaucoma}} & RETFound & $-.022$ & $-.008$ (19) & $-.0003$ (10) & $-.012$ & age, sex \\
 & DINOv2 & $-.008$ & $+.001$ (9) & $-.0009$ (10) & $-.001$ & age \\
 & Dual-IFM & $-.021$ & $-.003$ (13) & $+.0005$ (9) & $-.005$ & age, sex \\
\midrule
\multirow{3}{*}{\shortstack[l]{PAPILA\\glaucoma}} & RETFound & $-.057$ & $+.017$ (1) & -- & $+.010$ & age \\
 & DINOv2 & $-.090$ & $+.003$ (8) & -- & $-.003$ & age \\
 & Dual-IFM & $-.032$ & $+.010$ (1) & -- & $+.012$ & age \\
\midrule
\multirow{3}{*}{\shortstack[l]{CXR\\cardiomeg.}} & RAD-DINO & $-.016$ & $-.002$ (18) & $+.0005$ (5) & $-.003$ & age, sex \\
 & DINOv2-B & $-.026$ & $-.006$ (20) & $.0000$ (8) & $-.008$ & age, sex \\
 & DINOv2-L & $-.023$ & $-.008$ (20) & $-.0002$ (12) & $-.010$ & age, sex \\
\midrule
\multirow{3}{*}{\shortstack[l]{CXR\\effusion}} & RAD-DINO & $-.013$ & $.000$ (2) & $+.0004$ (1) & $-.001$ & age \\
 & DINOv2-B & $-.022$ & $.000$ (18) & $-.0004$ (17) & $-.001$ & age$^\dagger$ \\
 & DINOv2-L & $-.016$ & $.000$ (9) & $+.0003$ (2) & $-.001$ & age \\
\bottomrule
\end{tabular}}
\caption{\textbf{The matched test across 12 settings.} $\Delta$ all: erasure cost on all patients. $\bar E$: mean excess over the shuffled-demographic null after age matching or after joint age and sex matching, with the number of 20 matchings in which $E<0$. $\Delta_{\mathcal M}$: age-matched cost before subtracting the null. Joint matching was run for every setting except PAPILA, where 79 pairs are too few to split by sex. $^\dagger$Consistent but negligible residual ($|\bar E|<0.001$).}
\label{tab:main}
\end{table}

\begin{table}[t]
\centering\footnotesize\setlength{\tabcolsep}{2.5pt}
\adjustbox{max width=\columnwidth}{%
\begin{tabular}{@{}lccccccc@{}}
\toprule
 & Age+sex & \multicolumn{2}{c}{RETFound} & \multicolumn{2}{c}{DINOv2} & \multicolumn{2}{c}{Dual-IFM} \\
\cmidrule(lr){3-4}\cmidrule(lr){5-6}\cmidrule(l){7-8}
Label & AUC & AUC & Cost ($p$) & AUC & Cost ($p$) & AUC & Cost ($p$) \\
\midrule
Normal & .531 & .751 & $-.011$ (.010) & .765 & $-.011$ (.010) & .738 & $-.011$ (.010) \\
Diabetes & .568 & .742 & $-.012$ (.010) & .768 & $-.013$ (.010) & .726 & $-.012$ (.010) \\
Glaucoma & .632 & .827 & $-.022$ (.010) & .846 & $-.008$ (.040) & .821 & $-.021$ (.010) \\
Cataract & .725 & .973 & $-.010$ (.010) & .968 & $-.012$ (.010) & .972 & $-.011$ (.010) \\
AMD & .575 & .810 & $-.003$ (.80) & .854 & $-.003$ (.15) & .791 & $-.007$ (.089) \\
Hypertension & .525 & .757 & $-.005$ (.78) & .826 & $-.003$ (.44) & .782 & $+.005$ (.98) \\
Myopia & .582 & .986 & $-.004$ (.069) & .985 & $-.006$ (.050) & .988 & $-.006$ (.010) \\
\bottomrule
\end{tabular}}
\caption{\textbf{Erasure cost for all seven ODIR-5K labels.} Age+sex: AUC of a probe on age and sex alone. $p$: against 100 shuffled-demographic projections (minimum attainable 0.0099, shown as .010).}
\label{tab:odir}
\end{table}

\noindent\textbf{Erasure alone flags every setting.}
All six models encode sex and age (\cref{tab:enc}). Erasing these directions lowers disease AUC in all 12 settings, by more than erasing randomly reassigned demographics does (\cref{fig:main}a). On ODIR-5K, the glaucoma cost is $-0.022$ for RETFound, $-0.008$ for DINOv2, and $-0.021$ for Dual-IFM ($p\le0.04$), whereas removing two random orthonormal directions costs less than 0.001. The cost is not specific to glaucoma: normal, diabetes, glaucoma, and cataract all pay it in every model (\cref{tab:odir}). On PAPILA the costs are larger, at $-0.057$ $[-0.097,-0.023]$, $-0.090$ $[-0.133,-0.049]$, and $-0.032$ $[-0.065,-0.000]$ (all $p{=}0.0099$), and LEACE gives $-0.100$, $-0.124$, and $-0.054$. On chest radiographs every cost lies between $-0.013$ and $-0.026$, with intervals excluding zero, and LEACE agrees at every depth (\cref{tab:s_cxr}). Read at face value, every model relies on demographics everywhere.

\noindent\textbf{Age matching removes the PAPILA and effusion costs.}
On PAPILA, after matching 79 pairs, the cost disappears in all three models ($\Delta_{\mathcal M}$ from $-0.003$ to $+0.012$; single-matching $p\ge0.53$), and the excess is negative in only 1, 8, and 1 of 20 matchings (\cref{tab:main}). For RETFound, $-0.051$ of the $-0.057$ lies between age bands and only $-0.006$ within them (\cref{tab:s_bands}). Genuine glaucoma signal remains. On matched eyes, where age and sex alone predict at chance (AUC 0.44), the probes reach 0.72 to 0.80 across the 20 matchings. For effusion, a full cost of $-0.013$ to $-0.022$ shrinks by about 95\% after matching 3,033 pairs, and the excess is zero or negligible in every model.

\noindent\textbf{Residuals after age matching are sex shortcuts.}
In four settings, a small cost survives age matching in nearly every pairing, and an age-only test would call each of them an overlap. On ODIR-5K, RETFound keeps a cost in all 20 age matchings (mean $-0.012$) and exceeds the null in 19 of them ($\bar E{=}-0.008$), while Dual-IFM shows a weak trace ($-0.003$; 13 of 20) and DINOv2 none ($+0.001$; 9 of 20). For cardiomegaly, all three chest X-ray models keep an excess of $-0.002$ to $-0.008$ in 18 to 20 of 20 matchings. Matching on sex as well removes every one of these residuals without losing pairs (316 on ODIR-5K and 1,494 for cardiomegaly). The excess falls to between $-0.001$ and $+0.001$, negative in 5 to 12 of 20 matchings (\cref{tab:main}, \cref{fig:main}b). Chest X-ray features encode sex almost perfectly (AUC 0.99), so a sex shortcut is available to every model there. No setting shows an overlap within the resolution measured in \cref{sec:known}. Each erasure cost is a demographic shortcut: through age alone on PAPILA and for effusion, and through age and sex on ODIR-5K and for cardiomegaly.

\subsection{Validating the Test with Known Answers}
\label{sec:known}

\begin{table}[t]
\centering\footnotesize\setlength{\tabcolsep}{3pt}
\adjustbox{max width=\columnwidth}{%
\begin{tabular}{@{}llrr@{}}
\toprule
Experiment & Model & $\Delta$ all & $\bar E$ ($E{<}0$ of 10) \\
\midrule
\multicolumn{4}{@{}l}{\emph{Added age confounding, real labels (expected: $\Delta$ grows, $\bar E$ unchanged)}} \\
Random subsample (age gap 5.3\,y) & RETFound & $-.030$ & $-.006$ (9) \\
Younger controls (age gap 9.3\,y) & RETFound & $-.058$ & $-.005$ (10) \\
Random subsample (age gap 5.3\,y) & DINOv2 & $-.016$ & $+.002$ (3) \\
Younger controls (age gap 9.3\,y) & DINOv2 & $-.037$ & $-.002$ (7) \\
\midrule
\multicolumn{4}{@{}l}{\emph{Planted labels (expected: $\bar E\approx0$)}} \\
No demographic signal & RETFound & $-.002$ & $+.004$ (2) \\
 & DINOv2 & $-.001$ & $-.002$ (6) \\
Pure age shortcut & RETFound & $-.019$ & $.000$ (5) \\
 & DINOv2 & $-.018$ & $-.005$ (8)$^*$ \\
\midrule
\multicolumn{4}{@{}l}{\emph{Planted overlap in the features (expected: $\bar E<0$, growing with $\delta$)}} \\
$\delta=0$ & RETFound & $-.022$ & $+.001$ (4) \\
$\delta=0.25$ & RETFound & $-.081$ & $-.004$ (9) \\
$\delta=0.5$ & RETFound & $-.125$ & $-.007$ (10) \\
$\delta=0$ & DINOv2-L, CXR & $-.023$ & $-.000$ (6) \\
$\delta=0.25$ & DINOv2-L, CXR & $-.048$ & $-.010$ (10) \\
$\delta=0.5$ & DINOv2-L, CXR & $-.075$ & $-.024$ (10) \\
\bottomrule
\end{tabular}}
\caption{\textbf{Known-answer experiments} on ODIR-5K features, and on NIH cardiomegaly features for the CXR rows (10 matchings $\times$ 10 shuffles). Added confounding: all 316 glaucoma eyes plus 2,500 controls drawn at random or weighted toward younger ages. Planted labels: 10\% prevalence, driven by a feature direction with no demographic component, plus true age for the shortcut. Planted overlap: real labels, with diseased eyes shifted by $\delta$ standard deviations along the age direction, under joint age and sex matching. $^*$False positive; across six random draws, 3 in DINOv2 and none in RETFound (\cref{sec:supp_known}).}
\label{tab:known}
\end{table}

A test that finds no overlap is informative only if it would detect one that exists. We therefore built cases with known answers from real ODIR-5K features (\cref{tab:known}).

\noindent\textbf{Added confounding leaves the verdict unchanged.}
We kept all glaucoma eyes and drew 2,500 controls either at random or weighted toward younger patients, which widens the age gap from 5.3 to 9.3 years on the same features and disease. The full-data cost nearly doubles ($-0.030$ to $-0.058$ for RETFound, $-0.016$ to $-0.037$ for DINOv2), while the age-matched excess stays where it was ($-0.006$ to $-0.005$, and $+0.002$ to $-0.002$). Matching removes confounding that we introduced on purpose.

\noindent\textbf{A planted overlap is detected.}
We shifted the features of diseased patients along the age direction by $\delta$ standard deviations, so that the disease ``looks older'' in feature space whatever the patient's age. On ODIR-5K the excess moves from $+0.001$ at $\delta{=}0$ to $-0.004$ (9 of 10) at $\delta{=}0.25$ and $-0.007$ (10 of 10) at $\delta{=}0.5$. On chest radiographs the same plant in DINOv2-L cardiomegaly features gives $-0.010$ and $-0.024$ (10 of 10 each). The real cardiomegaly excess under joint matching ($-0.0002$) is therefore far below what even the smaller planted overlap produces. The test is conservative. At $\delta{=}0.5$ the cost on all patients is $-0.125$ but the excess is only $-0.007$, because the classifier rebuilds much of the erased signal from other, correlated directions. The test detects the part of an overlap that cannot be rebuilt this way. The plant lies along the erased direction, a best case, and the ODIR-5K detection at $\delta{=}0.25$ is within the resolution band.

\noindent\textbf{Resolution.}
We repeated the planted labels with six random disease directions per model. Labels with no demographic signal never met the overlap rule (0 of 12 draws). A planted pure age shortcut never met it in RETFound (0 of 6, mean excess $+0.001$) but met it in 3 of 6 draws in DINOv2, with false excesses up to $-0.005$ that exact age matching did not remove (\cref{sec:supp_known}). False positives are model-dependent and stay below 0.005 AUC, which we take as the test's resolution; every real excess under joint matching lies within 0.001 of zero.

\subsection{Ablation Studies}
\label{sec:ablation}

\begin{table}[t]
\centering\footnotesize\setlength{\tabcolsep}{2.2pt}
\adjustbox{max width=\columnwidth}{%
\begin{tabular}{@{}lccccc@{}}
\toprule
 & ODIR glauc. & PAPILA glauc. & CXR cardio. & CXR effus. & \\
Test variant & R\,D\,I & R\,D\,I & R\,B\,L & R\,B\,L & Flagged \\
\midrule
(a) Full-data cost $\Delta$ vs.\ null & $\bullet\,\bullet\,\bullet$ & $\bullet\,\bullet\,\bullet$ & $\bullet\,\bullet\,\bullet$ & $\bullet\,\bullet\,\bullet$ & 12/12 \\
(b) + age matching, CI only (no null) & $\bullet\,\circ\,\circ$ & $\circ\,\circ\,\circ$ & $\bullet\,\bullet\,\bullet$ & $\circ\,\bullet\,\bullet$ & 6/12 \\
(c) + null, one age matching & $\circ\,\circ\,\circ$ & $\circ\,\circ\,\circ$ & $\circ\,\bullet\,\bullet$ & $\circ\,\circ\,\circ$ & 2/12 \\
(d) + 20 age matchings, $\bar E$ & $\bullet\,\circ\,\odot$ & $\circ\,\circ\,\circ$ & $\bullet\,\bullet\,\bullet$ & $\circ\,\odot\,\circ$ & 4/12 \\
(e) + sex matching (ours) & $\circ\,\circ\,\circ$ & $\circ\,\circ\,\circ$ & $\circ\,\circ\,\circ$ & $\circ\,\odot\,\circ$ & 0/12 \\
\bottomrule
\end{tabular}}
\caption{\textbf{Ablating the test.} Settings flagged ($\bullet$) as each component is added. R: RETFound or RAD-DINO; D: DINOv2; I: Dual-IFM; B, L: DINOv2-B, DINOv2-L. $\odot$: weak (13/20) or negligible ($|\bar E|{<}0.001$) residual. Criteria: (a) $p{<}0.05$; (b) matched-cost CI excluding zero; (c) matched $p{<}0.05$; (d, e) the rule in \cref{sec:matched}. Row (e) adds exact sex matching wherever (d) left a residual.}
\label{tab:ablation}
\end{table}

\begin{table}[t]
\centering\footnotesize\setlength{\tabcolsep}{2.4pt}
\adjustbox{max width=\columnwidth}{%
\begin{tabular}{@{}llcccc@{}}
\toprule
 & & \multicolumn{4}{c}{Matched-set $p$ vs.\ shuffled null} \\
\cmidrule(l){3-6}
Setting & Model & main & $\tfrac13$ depth & final & MLP \\
\midrule
ODIR glauc. & RETFound & .079 & .40 & .81 & .39 \\
 & DINOv2 & .16 & .22 & .50 & .065 \\
 & Dual-IFM & .11 & .11 & .28 & .77 \\
PAPILA glauc. & RETFound & .82 & .82 & .83 & .58 \\
 & DINOv2 & .53 & .72 & .47 & .065 \\
 & Dual-IFM & .65 & .74 & .12 & .97 \\
CXR cardiomeg. & RAD-DINO & .12 & .21 & .16 & .52 \\
 & DINOv2-B & .030 & .020 & .069 & .032 \\
 & DINOv2-L & .020 & .56 & .050 & .16 \\
CXR effusion & RAD-DINO & .91 & .51 & .59 & .29 \\
 & DINOv2-B & .089 & .079 & .37 & .16 \\
 & DINOv2-L & .68 & .94 & .020 & .29 \\
\bottomrule
\end{tabular}}
\caption{\textbf{Robustness of the matched test.} $p$ of the matched cost against the shuffled null for one matching (100 nulls; 30 for the MLP probe), at the main layer, one-third depth, final depth, and with an MLP probe. \Cref{tab:s_rob,tab:s_cxr} report every cost.}
\label{tab:robust}
\end{table}

\noindent\textbf{Each component of the test changes verdicts.}
\Cref{tab:ablation} adds the components one at a time. Erasure alone (a) flags all 12 settings. Age matching without a null (b) clears the PAPILA shortcuts but still flags two effusion settings whose matched cost equals the generic cost of erasing two random directions. Adding the null with a single matching (c) misses residuals that are small but consistent, which averaging over 20 matchings (d) recovers. Matching on sex as well (e) shows that all four of those residuals are sex shortcuts, leaving no setting flagged.

\noindent\textbf{The null prevents false overlaps.}
At one-third depth, the matched cardiomegaly cost of DINOv2-L ($-0.023$) is no larger than its own shuffled null ($-0.024$; $p{=}0.56$). Early-layer features are dense enough that removing any two directions costs this much, and without the null this cost would be read as a strong overlap.

\noindent\textbf{Depth, probe, and label noise.}
The age-matched verdicts hold at one-third and final depth and with an MLP disease probe (\cref{tab:robust}). Every PAPILA and effusion check has matched $p\ge0.065$, except DINOv2-L effusion at the final layer ($-0.002$; $p{=}0.02$). On ODIR-5K, removing the 42 suspected-glaucoma eyes leaves RETFound's age-matched excess at $-0.006$ (17 of 20), so label noise does not create the residual.

\begin{figure*}[t]
\centering
\includegraphics[width=0.82\textwidth]{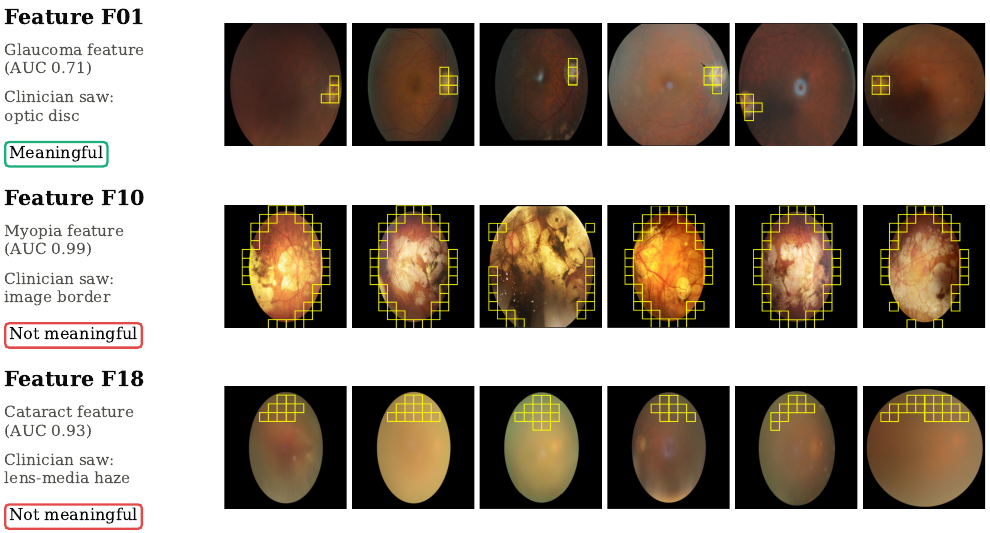}
\caption{\textbf{RETFound SAE features reviewed by a blinded ophthalmologist.} The six ODIR-5K images that activate each feature most; boxes mark patches at or above half of the maximum activation. F01 (glaucoma) responds to the optic disc and was judged clinically meaningful. F10 (myopia) and F18 (cataract) predict their labels well, respond to the image border and to lens haze, and were rejected.}
\label{fig:features}
\end{figure*}

\subsection{Fine-tuning Can Amplify a Shortcut}
\label{sec:finetune}

\begin{table}[t]
\centering\footnotesize\setlength{\tabcolsep}{2.2pt}
\adjustbox{max width=\columnwidth}{%
\begin{tabular}{@{}lccccc@{}}
\toprule
 & AUC & $\rho$ healthy & $\Delta\rho$ 95\% CI & AUC drop & $\Delta$drop 95\% CI \\
\midrule
\multicolumn{6}{@{}l}{\emph{PAPILA} (420 eyes)} \\
Frozen & .826 & .259 & -- & .042 & -- \\
FT seed 0 & .837 & .467 & [.118, .308] & .099 & [.006, .118] \\
FT seed 1 & .827 & .499 & [.142, .346] & .111 & [.015, .138] \\
FT seed 2 & .834 & .465 & [.110, .310] & .097 & [.003, .118] \\
\midrule
\multicolumn{6}{@{}l}{\emph{ODIR-5K} (6,713 eyes)} \\
Frozen & .827 & .107 & -- & .012 & -- \\
FT seed 0 & .910 & .150 & [.009, .071] & .011 & [$-$.022, .018] \\
FT seed 1 & .912 & .145 & [.006, .066] & .011 & [$-$.022, .017] \\
FT seed 2 & .912 & .152 & [.014, .073] & .011 & [$-$.021, .017] \\
\bottomrule
\end{tabular}}
\caption{\textbf{Partially fine-tuning RETFound.} $\rho$: Spearman correlation between glaucoma score and age in non-glaucoma eyes. Drop: AUC on all eyes minus AUC on age-matched eyes, averaged over 20 matchings. CIs: fine-tuned minus frozen, from 1,000 patient bootstraps.}
\label{tab:ft}
\end{table}

A deployed screening model would be fine-tuned, so we repeat the audit after partial fine-tuning with three seeds per dataset (\cref{tab:ft}). On PAPILA, fine-tuning leaves AUC unchanged (0.827 to 0.837 \vs 0.826). In every seed it nearly doubles the healthy-eye correlation $\rho$ and more than doubles the matched drop $D$ (\cref{eq:outside}), with all intervals excluding zero. The inside test on fine-tuned features again finds a shortcut: a full cost of $-0.040$ to $-0.041$, with intervals excluding zero, that vanishes after age matching ($p{=}0.57$ to $0.77$). On ODIR-5K, fine-tuning raises AUC from 0.827 to between 0.910 and 0.912 while $D$ stays at 0.011 against 0.012 when frozen, with intervals spanning zero in every seed. Here $\rho$ rises only from 0.11 to 0.15, and the age-matched cost on fine-tuned features ranges from $-0.002$ to $+0.001$ ($p{=}0.38$ to $0.46$). On the small, age-confounded dataset, fine-tuning strengthened the age shortcut. On the larger dataset, its gain held on age-matched eyes. The two datasets differ in both size and confounding, so these runs cannot attribute the contrast to either factor alone.

\subsection{Further Analysis}
\label{sec:analysis}

\begin{figure}[t]
\centering
\includegraphics[width=\columnwidth]{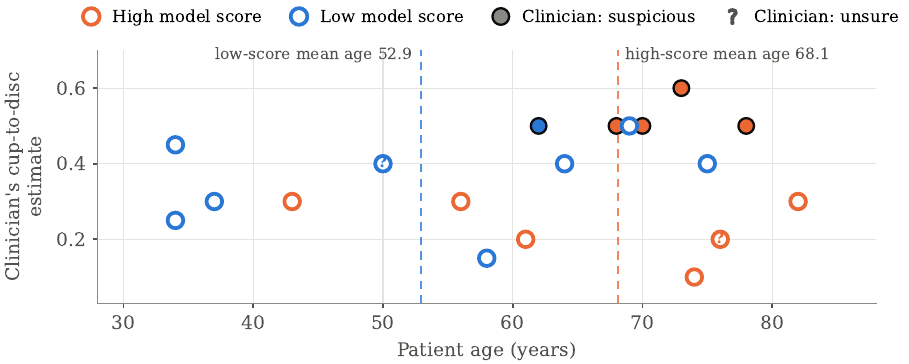}
\caption{\textbf{Healthy PAPILA eyes with the highest and lowest RETFound glaucoma scores}, reviewed without knowledge of age or score. High-score eyes are 15 years older ($p{=}0.03$) and have similar estimated cup-to-disc ratios ($p{=}0.87$): the score tracks age, and we detect no difference in cups ($n{=}10$ per group).}
\label{fig:eyes}
\end{figure}

\begin{table}[t]
\centering\footnotesize\setlength{\tabcolsep}{2pt}
\adjustbox{max width=\columnwidth}{%
\begin{tabular}{@{}llcccc@{}}
\toprule
 & & \multicolumn{2}{c}{Inside: encoded} & \multicolumn{2}{c}{Outside-in: healthy} \\
\cmidrule(lr){3-4}\cmidrule(l){5-6}
Setting & Model & Sex AUC & Age $R^2$ & $\rho$(score, age) & Score$\to$age${\ge}$60 \\
\midrule
ODIR glauc. & RETFound & .618 & .177 & .107 & .545 \\
 & DINOv2 & .619 & .106 & .074 & .531 \\
 & Dual-IFM & .610 & .082 & .103 & .544 \\
PAPILA glauc. & RETFound & .731 & .649 & .259 & .581 \\
 & DINOv2 & .705 & .337 & .298 & .634 \\
 & Dual-IFM & .664 & .483 & .270 & .613 \\
CXR cardiomeg. & RAD-DINO & .999 & .853 & .201 & .606 \\
 & DINOv2-B & .994 & .705 & .202 & .620 \\
 & DINOv2-L & .994 & .728 & .155 & .597 \\
CXR effusion & RAD-DINO & .999 & .848 & .271 & .647 \\
 & DINOv2-B & .993 & .684 & .262 & .657 \\
 & DINOv2-L & .994 & .713 & .243 & .628 \\
\bottomrule
\end{tabular}}
\caption{\textbf{Encoding versus leakage.} Left: sex AUC and age $R^2$ of linear probes on the main layer. Right: Spearman correlation of the disease score with age, and AUC of the score for age${\ge}60$, among patients without the disease.}
\label{tab:enc}
\end{table}

\begin{table}[t]
\centering\footnotesize\setlength{\tabcolsep}{3pt}
\adjustbox{max width=\columnwidth}{%
\begin{tabular}{@{}llcccccc@{}}
\toprule
 & & \multicolumn{3}{c}{Sex} & \multicolumn{3}{c}{Age ${\ge}$60} \\
\cmidrule(lr){3-5}\cmidrule(l){6-8}
Dataset & Model & Orig. & Proj. & LEACE & Orig. & Proj. & LEACE \\
\midrule
ODIR-5K & RETFound & .611 & .604 & .601 & .722 & .715 & .721 \\
 & DINOv2 & .632 & .595 & .599 & .709 & .689 & .687 \\
 & Dual-IFM & .615 & .598 & .592 & .693 & .675 & .679 \\
PAPILA & RETFound & .729 & .542 & .539 & .896 & .801 & .774 \\
 & DINOv2 & .668 & .447 & .427 & .835 & .610 & .634 \\
 & Dual-IFM & .639 & .482 & .571 & .841 & .598 & .689 \\
\bottomrule
\end{tabular}}
\caption{\textbf{Age survives linear erasure.} AUC of an MLP reading sex and age${\ge}60$ before erasure and after projection or LEACE. Linear probes on the erased features score 0.44--0.49.}
\label{tab:mlp}
\end{table}

\noindent\textbf{Encoding does not predict reliance.}
How much demographic information a model encodes did not predict how much its disease probe depends on it (\cref{tab:enc}). On PAPILA, DINOv2 encodes less age than RETFound ($R^2$ 0.34 \vs 0.65) yet pays the largest erasure cost ($-0.090$), and on ODIR-5K the same model pays the smallest. On chest radiographs, RAD-DINO encodes the most age ($R^2$ 0.85) and has the smallest cardiomegaly residual. Reliance depends on how the label relates to demographics in the data used to train the probe, which is what the matched test measures.

\noindent\textbf{The outside-in audit agrees.}
From scores alone, $\rho$ is about three times higher on PAPILA (0.26 to 0.30) than on ODIR-5K (0.07 to 0.11), and PAPILA scores predict age${\ge}60$ in healthy eyes (AUC 0.58 to 0.63). In a logistic model with age, sex, and score, age carries much more weight on PAPILA than on ODIR-5K (\cref{tab:s_out}). On chest radiographs, scores track age among No-Finding patients for both diseases ($\rho$ 0.16 to 0.27). Score-only checks therefore flag demographic dependence, but only the matched test identifies its source.

\noindent\textbf{Linear erasure hides age without removing it.}
After either erasure, linear probes read sex and age${\ge}60$ at chance (AUC 0.44 to 0.49), the usual criterion for success. A one-hidden-layer MLP still reads age (\cref{tab:mlp}). For RETFound on ODIR-5K it reaches 0.715 after projection and 0.721 after LEACE, against 0.722 before erasure, and on PAPILA age remains readable in every model (0.60 to 0.80). The LEACE guarantee covers linear classifiers only~\cite{belrose2023leace,ravfogel2022kernel}, so erased features can still supply age to a non-linear downstream model.

\subsection{Visualization}
\label{sec:vis}

\noindent\textbf{Features that mix disease and age.}
We trained top-$k$ sparse autoencoders (SAEs)~\cite{gao2024sae}, which rewrite features as a few active units (8,192 units, 48 active per patch), on ODIR-5K patch tokens. Among each model's 20 SAE units most predictive of glaucoma, 7 (RETFound), 6 (Dual-IFM), and 2 (DINOv2) also rank among its 20 most predictive of age${\ge}60$, against a chance level of at most 0.3 (\cref{tab:s_sae}). Single units therefore mix glaucoma and age even though the matched test finds no reliance beyond demographics, so mixing at the level of units does not establish an overlap. A practicing ophthalmologist, blinded to labels, judged 13 of 19 RETFound features clinically meaningful (\cref{fig:features}). Four of the six rejected were border or image-quality artifacts.

\noindent\textbf{What high scores track.}
Among healthy PAPILA eyes, the 10 with the highest glaucoma scores were 15 years older than the 10 with the lowest, yet their estimated cup-to-disc ratios were similar (\cref{fig:eyes}). 
\section{Conclusion}
\label{sec:conclusion}
A drop in accuracy after demographic erasure does not tell an auditor whether a medical foundation model uses a shortcut or encodes shared biology. The matched, null-calibrated test introduced here makes that distinction from frozen features, and known-answer experiments show that it ignores added confounding and detects a planted overlap. Across six retinal and six chest X-ray settings, no erasure cost we measured exceeds the test's resolution after age and sex matching, so each is best explained as a demographic shortcut. Matching on age alone would have attributed four sex shortcuts to biology, so audits should match on every recorded demographic that the erasure removes. Debiasing should also be verified with a probe as flexible as the downstream model, and models re-audited after fine-tuning, which on a small confounded dataset deepened the shortcut at unchanged AUC.

\noindent\textbf{Limitations.}
The test resolves excesses of about 0.005 AUC: a strongly age-driven planted shortcut produced false excesses up to that size in 3 of 6 draws in one of two models, and planted overlaps are attenuated because probes rebuild erased signal from correlated directions. Smaller overlaps may therefore go undetected. PAPILA is small (79 pairs, matched on age only), and the ODIR-5K and NIH labels are mined from text. Matching does not control acquisition site or device. RAD-DINO was pretrained without labels on a corpus that includes NIH ChestX-ray14. Fine-tuning used a single recipe, the clinician review is a single-reader pilot, and neither retinal dataset records ethnicity.
\section*{Acknowledgments}
I thank Professor Aaditeshwar Seth and Sarvagya Upadhyay for their mentorship and guidance throughout this work, and Dr.\ Pooja Godara for reviewing the model features and fundus photographs as the study's ophthalmologist.

\section*{LLM Usage Statement}
We used large language models (Claude and Gemini) as assistants throughout this work. They helped write and debug the analysis code (feature extraction, probing, erasure, matching, null, and fine-tuning scripts), proposed robustness analyses, helped interpret intermediate outputs, and helped draft and edit the text and LaTeX. The author designed the study, ran every experiment, checked each reported number against the experiment outputs, verified every reference against its published source, and takes full responsibility for the content. No LLM output was used as data.
{
    \small
    \bibliographystyle{ieeenat_fullname}
    \bibliography{main}
}

\input{appendix}

\end{document}

%% file: appendix.tex
\clearpage
\onecolumn
\appendix
\setcounter{table}{0}
\setcounter{figure}{0}
\renewcommand{\thesection}{\Alph{section}}
\renewcommand{\thetable}{A\arabic{table}}
\renewcommand{\thefigure}{A\arabic{figure}}
\begin{center}{\Large\bf Appendix}\end{center}
\vspace{6pt}

\noindent This appendix gives the details behind the main text. Tables and figures numbered with the prefix A belong to the appendix.

\section{Code and Data Availability}
\label{sec:supp_code}
\noindent The code is available at \url{https://github.com/StellarDrift15/Retinal_Shortcuts} under the Apache 2.0 license. It contains the six audit notebooks (three foundation models on each retinal dataset), the chest X-ray notebook, and the preliminary experiments, with every output saved in the notebooks. Fine-tuning runs inside the RETFound notebooks, and the README maps each table and figure to the cells that produce it. All three datasets are public: ODIR-5K and PAPILA are available from their original releases, and NIH ChestX-ray14 is available from the National Institutes of Health. The notebooks download the data and model weights, so no patient data are redistributed.

\section{Terms Used in This Appendix}
\label{sec:supp_terms}
\noindent The appendix uses the following terms, each defined as in the main text.
\begin{itemize}[leftmargin=*,itemsep=2pt,topsep=2pt]
\item \textbf{Feature vector and direction.} A foundation model turns each image into a list of numbers, its feature vector. A property such as age corresponds to a direction in this space: moving along it makes the image look, to the model, older or younger.
\item \textbf{Probe.} A simple classifier (logistic regression) trained on frozen feature vectors to predict a label such as glaucoma. It tells us what the features contain without changing the model.
\item \textbf{AUC.} Area under the ROC curve: 0.5 is chance and 1.0 is perfect. Costs are reported as changes in AUC.
\item \textbf{Erasure.} Removing the age and sex directions from every feature vector, then retraining the probe. We use a projection along the two directions (mean difference) and LEACE, a method that provably prevents any linear classifier from recovering age and sex. The \textbf{cost} is the resulting change in disease AUC.
\item \textbf{Shortcut and overlap.} A shortcut means the probe uses age or sex because they happen to predict the disease in this dataset. An overlap means the disease itself changes the image along the same directions as age or sex.
\item \textbf{Matching.} Pairing each diseased patient with a healthy patient of similar age (within 2 years; the \textbf{caliper}), or of the same sex and similar age (\textbf{joint matching}). Within matched pairs, age and sex no longer predict the disease.
\item \textbf{Shuffled-demographic null.} The same erasure, but with age and sex randomly reassigned between patients. Its cost is what removing two unrelated directions costs by chance.
\item \textbf{Excess cost} ($E$, mean $\bar E$). The cost on matched patients minus the null's cost on the same patients. A shortcut gives an excess near zero; an overlap gives a negative excess. We repeat the matching 20 times and count how many repeats have $E<0$ ($n_-$).
\item \textbf{Folds and bootstrap.} Results are cross-validated in 5 folds that keep all images of a patient together. Confidence intervals come from resampling patients 1,000 times.
\item \textbf{MLP.} A small neural network (one hidden layer) used as a more flexible probe than logistic regression.
\item \textbf{SAE.} A sparse autoencoder, which rewrites each feature vector as a few active units that are easier to inspect individually.
\end{itemize}

\section{Implementation Details}
\label{sec:supp_impl}
\noindent This section lists every setting needed to reproduce the results.

\noindent\textbf{Preprocessing and features.}
Retinal images were resized to a square without cropping. RETFound and DINOv2 received $224{\times}224$ inputs with ImageNet normalization. Dual-IFM received $256{\times}256$ inputs, its pretraining resolution, with its own normalization (mean 0.397, 0.233, 0.131; standard deviation 0.309, 0.194, 0.143). Chest radiographs passed through each model's own Hugging Face image processor ($518{\times}518$ for RAD-DINO; resize to 256 and center crop to 224 for DINOv2). An image feature is the mean of the patch tokens, or the spatial mean for convolutional stages, at one-third, two-thirds, and full depth: blocks 8, 16, and 24 for ViT-L; blocks 4, 8, and 12 for ViT-B; and the residual stages \texttt{layer2}, \texttt{layer3}, and \texttt{layer4} of Dual-IFM's ResNet-50 backbone. Chest X-ray features were extracted in bfloat16.

\noindent\textbf{Probes.}
Disease, sex, and age${\ge}60$ probes are logistic regressions on standardized features ($C{=}1$, L-BFGS). Continuous age is predicted by ridge regression ($\alpha{=}10$) and reported as out-of-fold $R^2$. The layer sweep and the probes on SAE features use $C{=}0.1$. The MLP probe has one hidden layer of 256 units, an L2 penalty of $10^{-2}$, early stopping, and at most 500 iterations.

\noindent\textbf{Erasure.}
The mean-difference projection computes the training-fold mean differences for sex and for age${\ge}60$, orthogonalizes them by Gram--Schmidt, and projects both out of the training and test features. LEACE uses the \texttt{concept-erasure} package with sex and standardized age as targets and is fitted on the training fold only.

\noindent\textbf{Folds, matching and nulls.}
Patient-grouped 5-fold splits were computed once per dataset, saved, and reloaded for every model and method, except in the robustness checks of the RETFound and Dual-IFM ODIR-5K notebooks (other depths, the MLP probe, and the exclusion of suspected glaucoma; \cref{tab:s_rob}), which build their own patient-grouped 5-fold split. Rerunning the main RETFound glaucoma test on that split gives a full cost of $-0.019$ [$-0.027$, $-0.011$] and a matched cost of $-0.003$ ($p{=}0.49$), the same verdict as the main analysis; matched sets use new grouped folds. Matching is greedy, without replacement, in random order, with a 2-year caliper. The matching paired all 316 ODIR-5K glaucoma eyes, all 307 cataract eyes, 79 of the 87 PAPILA glaucoma eyes, and every chest X-ray positive. The shuffled null gives each patient the sex and age of another, randomly chosen patient and rebuilds the erased directions in exactly the same way. The excess statistic uses 20 matchings (seeds 0 to 19) with 20 shuffles each.

\noindent\textbf{Fine-tuning.}
We trained the last 4 of RETFound's 24 blocks, its final norm, and a new linear head, keeping the remaining weights frozen. The image feature is the mean of the final-layer patch tokens. Optimization used AdamW (learning rate $10^{-5}$ for the backbone and $10^{-3}$ for the head, weight decay 0.05), a one-cycle schedule with 10\% warm-up, batch size 16, bfloat16 autocast, and binary cross-entropy with positive-class weighting. Training ran for a fixed 15 epochs on PAPILA and 8 on ODIR-5K with no early stopping, so test folds never influenced training. Augmentation consisted of resizing to 236, random cropping to 224, horizontal flips, rotations of up to $10^\circ$, and color jitter (0.1, 0.1). The seed for each run was $42+100\cdot\text{run}+\text{fold}$. For each outer fold, we re-extracted features for all images with that fold's model and trained the inside-test probes on that fold's training images only. Each seed took under one hour on a single NVIDIA L4 or A100 GPU.

\noindent\textbf{Sparse autoencoders.}
We trained top-$k$ SAEs with 8,192 features and $k{=}48$ on individual patch tokens of the main layer of each model on ODIR-5K, holding out 10\% of patients. Training used 30,000 Adam steps (learning rate $3{\times}10^{-4}$, batch size 4,096), centered and scaled inputs, a unit-norm decoder initialized from the encoder, and re-initialization of dead features every 4,000 steps. The value of a feature for an image is its maximum over patches.

\section{Datasets and Labels}
\label{sec:supp_data}
\noindent This section explains which images we used and how each diagnosis label was assigned.

\noindent\textbf{ODIR-5K.} The patient records yield 6,716 eyes. We removed three (a missing file or a non-fundus image), leaving 6,713 eyes from 3,358 patients. Labels come from explicit rules on the per-eye diagnostic keywords. An eye counts as normal only if its sole diagnostic keyword is ``normal fundus,'' with image-quality notes allowed. Aggregated to patients, our labels agree exactly with the ODIR-5K patient-level columns for all seven labels. The glaucoma label comprises 274 confirmed and 42 suspected cases.

\noindent\textbf{PAPILA.} The dataset contains 488 images of 244 patients. Excluding 68 glaucoma-suspect eyes leaves 420 eyes of 210 patients (87 glaucoma, 333 healthy), and the diagnosis counts match the published ones.

\noindent\textbf{NIH ChestX-ray14.} We used a public 224-pixel release that includes patient identifier, age, sex, and labels, and dropped images with recorded ages outside 18 to 95. For each patient we kept one image: a cardiomegaly image if one existed, otherwise an effusion image, otherwise a No-Finding image. Patients with only other findings were dropped. We kept all 1,496 cardiomegaly patients and sampled, with a fixed seed, 2,500 effusion-only and 8,000 No-Finding patients, for 11,996 images in total. The effusion cohort has 3,033 positives: the 2,500 effusion-only patients and 533 cardiomegaly patients whose image also shows effusion. With one image per patient, patient-level and image-level bootstraps coincide.

\section{Full Retinal Results}
\label{sec:supp_retina}
\noindent These tables give every retinal number behind the main text's summaries. In short, the age-matched cost stays near zero under every robustness check on PAPILA, and the ODIR-5K residual is small and consistent.

\Cref{tab:odir} reports the erasure costs for all seven ODIR-5K labels, and \cref{tab:s_bands} splits the glaucoma cost by age band.

\begin{table}[H]
\centering\footnotesize\setlength{\tabcolsep}{2pt}
\adjustbox{max width=\textwidth}{%
\begin{tabular}{@{}llrrrr@{}}
\toprule
Data & Model & Full & Within bands [95\% CI] & Between & Matched ($p$) \\
\midrule
ODIR-5K & RETFound & $-.022$ & $-.018$ [$-.025$, $-.011$] & $-.003$ & $-.016$ (.079) \\
 & DINOv2 & $-.008$ & $-.004$ [$-.010$, $.001$] & $-.003$ & $-.004$ (.16) \\
 & Dual-IFM & $-.021$ & $-.016$ [$-.024$, $-.008$] & $-.004$ & $-.006$ (.11) \\
PAPILA & RETFound & $-.057$ & $-.006$ [$-.042$, $.030$] & $-.051$ & $+.002$ (.82) \\
 & DINOv2 & $-.090$ & $-.034$ [$-.074$, $.006$] & $-.056$ & $-.006$ (.53) \\
 & Dual-IFM & $-.032$ & $+.022$ [$-.009$, $.056$] & $-.054$ & $+.006$ (.65) \\
\bottomrule
\end{tabular}}
\caption{\textbf{Where the glaucoma erasure cost lies.} Within: cost inside the age bands $<$60, 60--69, and $\ge$70, weighted by positives. Between: the remainder. Matched: one primary matching with 100 nulls. A shortcut places the cost between bands, and an overlap places it within them.}
\label{tab:s_bands}
\end{table}

\begin{table}[H]
\centering\footnotesize\setlength{\tabcolsep}{3pt}
\adjustbox{max width=\textwidth}{%
\begin{tabular}{@{}lllccccc@{}}
\toprule
Dataset, label & Model & Check & AUC & Full cost [95\% CI] & Pairs & $\Delta_{\mathcal M}$ [95\% CI] & $p$ \\
\midrule
ODIR-5K, glaucoma & RETFound & no suspects, layer 16 & .8199 & $-.0233$ [$-.0317$, $-.0152$] & 274 & $+.0029$ [$-.0046$, $.0109$] & .951 \\
 &  & layer 8 & .7923 & $-.0260$ [$-.0367$, $-.0155$] & 316 & $-.0081$ [$-.0177$, $.0027$] & .396 \\
 &  & layer 24 & .8273 & $-.0117$ [$-.0196$, $-.0028$] & 316 & $-.0016$ [$-.0089$, $.0051$] & .812 \\
 &  & MLP probe & .8384 & $+.0040$ [$-.0157$, $.0197$] & 316 & $+.0252$ [$.0052$, $.0497$] & .387 \\
 & DINOv2 & no suspects, layer 16 & .8461 & $-.0029$ [$-.0073$, $.0015$] & 274 & $-.0005$ [$-.0075$, $.0066$] & .228 \\
 &  & layer 8 & .8263 & $-.0390$ [$-.0515$, $-.0270$] & 316 & $-.0099$ [$-.0200$, $.0025$] & .218 \\
 &  & layer 24 & .8142 & $-.0080$ [$-.0131$, $-.0032$] & 316 & $-.0011$ [$-.0067$, $.0051$] & .495 \\
 &  & MLP probe & .8456 & $-.0128$ [$-.0311$, $.0052$] & 316 & $-.0157$ [$-.0347$, $.0042$] & .065 \\
 & Dual-IFM & no suspects, \texttt{layer3} & .8339 & $-.0196$ [$-.0311$, $-.0092$] & 274 & $-.0018$ [$-.0105$, $.0070$] & .574 \\
 &  & \texttt{layer2} & .8270 & $-.0174$ [$-.0264$, $-.0084$] & 316 & $-.0058$ [$-.0153$, $.0036$] & .109 \\
 &  & \texttt{layer4} & .8140 & $-.0030$ [$-.0096$, $.0032$] & 316 & $+.0018$ [$-.0040$, $.0086$] & .277 \\
 &  & MLP probe & .8206 & $-.0040$ [$-.0275$, $.0191$] & 316 & $+.0032$ [$-.0177$, $.0242$] & .774 \\
\midrule
ODIR-5K, cataract & RETFound & layer 8 & .9719 & $-.0095$ [$-.0150$, $-.0053$] & 307 & $-.0058$ [$-.0130$, $-.0003$] & .139 \\
 &  & layer 24 & .9770 & $-.0083$ [$-.0179$, $-.0023$] & 307 & $-.0004$ [$-.0046$, $.0037$] & .515 \\
 &  & MLP probe & .9715 & $-.0126$ [$-.0226$, $-.0026$] & 307 & $+.0034$ [$-.0068$, $.0144$] & .871 \\
 & DINOv2 & layer 8 & .9702 & $-.0108$ [$-.0186$, $-.0050$] & 307 & $+.0008$ [$-.0048$, $.0065$] & .733 \\
 &  & layer 24 & .9720 & $-.0126$ [$-.0213$, $-.0063$] & 307 & $+.0004$ [$-.0029$, $.0034$] & .713 \\
 &  & MLP probe & .9691 & $-.0210$ [$-.0330$, $-.0110$] & 307 & $-.0079$ [$-.0158$, $-.0005$] & .097 \\
 & Dual-IFM & \texttt{layer2} & .9717 & $-.0145$ [$-.0224$, $-.0081$] & 307 & $-.0039$ [$-.0111$, $.0034$] & .188 \\
 &  & \texttt{layer4} & .9656 & $-.0102$ [$-.0170$, $-.0053$] & 307 & $+.0006$ [$-.0033$, $.0041$] & .495 \\
 &  & MLP probe & .9749 & $-.0075$ [$-.0167$, $-.0007$] & 307 & $-.0024$ [$-.0107$, $.0054$] & .645 \\
\midrule
PAPILA, glaucoma & RETFound & layer 8 & .7747 & $-.0845$ [$-.1249$, $-.0476$] & 79 & $+.0187$ [$-.0189$, $.0566$] & .822 \\
 &  & layer 24 & .8604 & $-.0438$ [$-.0740$, $-.0138$] & 79 & $+.0051$ [$-.0072$, $.0176$] & .832 \\
 &  & MLP probe & .7686 & $-.0337$ [$-.0966$, $.0258$] & 79 & $+.0691$ [$.0001$, $.1442$] & .581 \\
 & DINOv2 & layer 8 & .8020 & $-.0994$ [$-.1384$, $-.0610$] & 79 & $-.0013$ [$-.0330$, $.0291$] & .723 \\
 &  & layer 24 & .8054 & $-.0570$ [$-.1002$, $-.0188$] & 79 & $-.0050$ [$-.0161$, $.0055$] & .465 \\
 &  & MLP probe & .8027 & $-.0703$ [$-.1181$, $-.0228$] & 79 & $-.0417$ [$-.1085$, $.0249$] & .065 \\
 & Dual-IFM & \texttt{layer2} & .8138 & $-.0455$ [$-.0753$, $-.0192$] & 79 & $+.0119$ [$-.0043$, $.0296$] & .743 \\
 &  & \texttt{layer4} & .8010 & $-.0694$ [$-.1186$, $-.0177$] & 79 & $-.0189$ [$-.0383$, $.0007$] & .119 \\
 &  & MLP probe & .7843 & $-.0659$ [$-.1265$, $-.0113$] & 79 & $+.0663$ [$.0080$, $.1362$] & .968 \\
\bottomrule
\end{tabular}}
\caption{\textbf{Glaucoma and cataract robustness checks, with intervals.} Full-data cost on all eyes and matched cost $\Delta_{\mathcal M}$ after retraining on one primary age matching, with 95\% patient-bootstrap intervals, and $p$ against the shuffled null (100 nulls for linear probes, 30 for the MLP probe). Main layers: 16 for RETFound and DINOv2, \texttt{layer3} for Dual-IFM. On PAPILA the MLP disease probe is weak on matched eyes (AUC 0.57 to 0.64), so its rows are uninformative.}
\label{tab:s_rob}
\end{table}

\begin{table}[H]
\centering\footnotesize\setlength{\tabcolsep}{3pt}
\begin{tabular}{@{}llccc@{}}
\toprule
Dataset & Model & $\Delta_{\mathcal M}$ mean (range) & $\bar E$ (range) & $E{<}0$ \\
\midrule
ODIR-5K & RETFound & $-.012$ ($-.021$, $-.004$) & $-.008$ & 19/20 \\
 & \quad no suspects & -- & $-.006$ & 17/20 \\
 & DINOv2 & $-.001$ & $+.001$ & 9/20 \\
 & \quad no suspects & -- & $-.001$ & 9/20 \\
 & Dual-IFM & $-.005$ & $-.003$ & 13/20 \\
 & \quad no suspects & -- & $-.0035$ & 16/20 \\
PAPILA & RETFound & $+.010$ & $+.017$ ($-.006$, $+.041$) & 1/20 \\
 & DINOv2 & $-.003$ & $+.003$ ($-.007$, $+.023$) & 8/20 \\
 & Dual-IFM & $+.012$ & $+.010$ ($-.007$, $+.022$) & 1/20 \\
\bottomrule
\end{tabular}
\caption{\textbf{Glaucoma results over 20 random matchings} (20 shuffles each). On ODIR-5K, $\Delta_{\mathcal M}$ is negative in 20 of 20 matchings for RETFound, 16 for Dual-IFM, and 14 for DINOv2. On PAPILA, matched glaucoma AUC stays between 0.72 and 0.80 across matchings (RETFound 0.762 to 0.774, DINOv2 0.739 to 0.796, Dual-IFM 0.723 to 0.771), while age and sex alone predict at chance (AUC 0.44).}
\label{tab:s_20}
\end{table}

\noindent\textbf{Cataract: an age-related disease without a shortcut.} Cataract is strongly age-related in ODIR-5K (age and sex alone give AUC 0.725), yet the models did not depend on age to detect it. On 307 age-matched pairs, demographics fell to chance (AUC 0.473) while the models stayed at 0.955 to 0.961, and retraining on matched eyes left the erasure cost at essentially zero for every model ($p \ge 0.60$). The test therefore does not flag a disease merely because it is associated with age.

\noindent\textbf{Binary versus continuous age.} On PAPILA, projection also leaves part of continuous age readable even to a linear probe ($R^2$ 0.19, 0.09, and 0.15 for RETFound, DINOv2, and Dual-IFM, down from 0.65, 0.34, and 0.48), because it removes only the binary age${\ge}60$ direction; LEACE, which targets continuous age, brings this to zero.

\noindent\textbf{Age encoding by depth.} On ODIR-5K, a linear probe reads continuous age about equally well at one-third, two-thirds, and full depth in RETFound ($R^2$ 0.162, 0.177, 0.184), while it fades with depth in DINOv2 (0.142, 0.106, 0.065) and disappears by the final stage in Dual-IFM (0.144, 0.082, $-0.215$; \texttt{layer2} to \texttt{layer4}). How much age a layer encodes therefore varies by model and depth, which is consistent with the main finding that encoding does not predict reliance (\cref{tab:enc}).

\section{Full Chest X-ray Results}
\label{sec:supp_cxr}
\noindent These tables give every chest X-ray number. Erasure costs AUC at every depth and with both erasure methods, and most of that cost disappears after age matching.

\begin{table}[H]
\centering\footnotesize\setlength{\tabcolsep}{2.5pt}
\adjustbox{max width=\textwidth}{%
\begin{tabular}{@{}lllccccccc@{}}
\toprule
Model & Check & Disease & AUC & Full cost [95\% CI] & Matched AUC & Matched cost [95\% CI] & Null mean & $p$ & LEACE cost [95\% CI] \\
\midrule
RAD-DINO & main (L8) & cardiomeg. & .928 & $-.0161$ [$-.0194$, $-.0129$] & .915 & $-.0030$ [$-.0047$, $-.0012$] & $-.0017$ & .119 & $-.0125$ [$-.0151$, $-.0099$] \\
 & & effusion & .936 & $-.0133$ [$-.0160$, $-.0106$] & .924 & $-.0003$ [$-.0007$, $.0001$] & $-.0008$ & .911 & $-.0119$ [$-.0143$, $-.0096$] \\
DINOv2-B & main (L8) & cardiomeg. & .879 & $-.0259$ [$-.0304$, $-.0215$] & .828 & $-.0061$ [$-.0095$, $-.0026$] & $-.0018$ & .030 & $-.0204$ [$-.0245$, $-.0165$] \\
 & & effusion & .911 & $-.0218$ [$-.0251$, $-.0186$] & .892 & $-.0016$ [$-.0024$, $-.0009$] & $-.0006$ & .089 & $-.0181$ [$-.0211$, $-.0152$] \\
DINOv2-L & main (L16) & cardiomeg. & .854 & $-.0226$ [$-.0269$, $-.0182$] & .831 & $-.0063$ [$-.0095$, $-.0031$] & $-.0015$ & .020 & $-.0168$ [$-.0201$, $-.0132$] \\
 & & effusion & .903 & $-.0164$ [$-.0196$, $-.0136$] & .873 & $-.0006$ [$-.0011$, $-.0001$] & $-.0008$ & .683 & $-.0143$ [$-.0171$, $-.0117$] \\
\midrule
RAD-DINO & L4 & cardiomeg. & .912 & $-.0175$ [$-.0211$, $-.0142$] & .897 & $-.0027$ [$-.0048$, $-.0006$] & $-.0018$ & .208 & $-.0156$ [$-.0189$, $-.0124$] \\
 & & effusion & .917 & $-.0180$ [$-.0211$, $-.0150$] & .906 & $-.0008$ [$-.0011$, $-.0004$] & $-.0009$ & .505 & $-.0183$ [$-.0213$, $-.0154$] \\
 & L12 & cardiomeg. & .920 & $-.0166$ [$-.0200$, $-.0131$] & .905 & $-.0033$ [$-.0053$, $-.0013$] & $-.0019$ & .158 & $-.0128$ [$-.0158$, $-.0100$] \\
 & & effusion & .936 & $-.0127$ [$-.0153$, $-.0101$] & .925 & $-.0007$ [$-.0009$, $-.0003$] & $-.0008$ & .594 & $-.0109$ [$-.0132$, $-.0088$] \\
DINOv2-B & L4 & cardiomeg. & .861 & $-.0277$ [$-.0319$, $-.0231$] & .813 & $-.0062$ [$-.0095$, $-.0029$] & $-.0016$ & .020 & $-.0226$ [$-.0266$, $-.0184$] \\
 & & effusion & .898 & $-.0230$ [$-.0267$, $-.0194$] & .879 & $-.0017$ [$-.0026$, $-.0007$] & $-.0006$ & .079 & $-.0182$ [$-.0214$, $-.0150$] \\
 & L12 & cardiomeg. & .858 & $-.0256$ [$-.0301$, $-.0210$] & .830 & $-.0056$ [$-.0087$, $-.0026$] & $-.0020$ & .069 & $-.0198$ [$-.0237$, $-.0160$] \\
 & & effusion & .898 & $-.0195$ [$-.0229$, $-.0163$] & .879 & $-.0006$ [$-.0012$, $-.0000$] & $-.0007$ & .366 & $-.0155$ [$-.0183$, $-.0125$] \\
DINOv2-L & L8 & cardiomeg. & .845 & $-.0367$ [$-.0419$, $-.0317$] & .829 & $-.0232$ [$-.0282$, $-.0181$] & $-.0236$ & .564 & $-.0262$ [$-.0306$, $-.0217$] \\
 & & effusion & .895 & $-.0285$ [$-.0321$, $-.0250$] & .879 & $-.0073$ [$-.0089$, $-.0059$] & $-.0121$ & .941 & $-.0220$ [$-.0251$, $-.0189$] \\
 & L24 & cardiomeg. & .855 & $-.0271$ [$-.0317$, $-.0224$] & .825 & $-.0047$ [$-.0078$, $-.0016$] & $-.0018$ & .050 & $-.0224$ [$-.0263$, $-.0183$] \\
 & & effusion & .894 & $-.0175$ [$-.0205$, $-.0146$] & .872 & $-.0019$ [$-.0024$, $-.0014$] & $-.0009$ & .020 & $-.0142$ [$-.0170$, $-.0116$] \\
\midrule
RAD-DINO & MLP (L8) & cardiomeg. & .926 & $-.0047$ [$-.0094$, $-.0003$] & .912 & $+.0081$ [$.0034$, $.0133$] & $+.0076$ & .516 & $-.0075$ [$-.0122$, $-.0027$] \\
 & & effusion & .937 & $-.0055$ [$-.0085$, $-.0021$] & .932 & $-.0003$ [$-.0031$, $.0026$] & $+.0008$ & .290 & $-.0054$ [$-.0084$, $-.0023$] \\
DINOv2-B & MLP (L8) & cardiomeg. & .879 & $-.0145$ [$-.0193$, $-.0098$] & .862 & $-.0106$ [$-.0167$, $-.0045$] & $-.0023$ & .032 & $-.0130$ [$-.0174$, $-.0086$] \\
 & & effusion & .910 & $-.0203$ [$-.0251$, $-.0160$] & .901 & $-.0006$ [$-.0033$, $.0020$] & $+.0013$ & .161 & $-.0123$ [$-.0160$, $-.0083$] \\
DINOv2-L & MLP (L16) & cardiomeg. & .861 & $-.0074$ [$-.0140$, $-.0008$] & .859 & $-.0034$ [$-.0082$, $.0019$] & $-.0005$ & .161 & $+.0009$ [$-.0053$, $.0067$] \\
 & & effusion & .906 & $-.0110$ [$-.0149$, $-.0072$] & .904 & $-.0020$ [$-.0054$, $.0012$] & $-.0013$ & .290 & $-.0111$ [$-.0145$, $-.0075$] \\
\bottomrule
\end{tabular}}
\caption{\textbf{All chest X-ray results} (NIH ChestX-ray14, one image per patient; cardiomegaly $n{=}9{,}496$ with 1,496 matched pairs; effusion $n{=}11{,}033$ with 3,033 pairs). Matched results use one primary matching with 100 nulls (30 for the MLP probe). Mean age is 50.9 for cardiomegaly and 51.4 for effusion positives, against 46.3 for negatives. LEACE cost: full-data cost of LEACE on sex and standardized age, followed by the row's own probe (linear or MLP).}
\label{tab:s_cxr}
\end{table}

\begin{table}[H]
\centering\footnotesize\setlength{\tabcolsep}{3pt}
\begin{tabular}{@{}llcccc@{}}
\toprule
Model & Disease & Real $\Delta_{\mathcal M}$ & Shuffled & $\bar E$ (range) & $E{<}0$ \\
\midrule
RAD-DINO & cardiomeg. & $-.0034$ & $-.0017$ & $-.0017$ ($-.0040$, $+.0005$) & 18/20 \\
 & effusion & $-.0006$ & $-.0009$ & $+.0003$ ($-.0001$, $+.0007$) & 2/20 \\
DINOv2-B & cardiomeg. & $-.0076$ & $-.0016$ & $-.0059$ ($-.0103$, $-.0034$) & 20/20 \\
 & effusion & $-.0011$ & $-.0007$ & $-.0004$ ($-.0009$, $+.0002$) & 18/20 \\
DINOv2-L & cardiomeg. & $-.0096$ & $-.0016$ & $-.0080$ ($-.0164$, $-.0024$) & 20/20 \\
 & effusion & $-.0008$ & $-.0008$ & $+.0000$ ($-.0010$, $+.0005$) & 9/20 \\
\bottomrule
\end{tabular}
\caption{\textbf{Chest X-ray excess over 20 matchings} (20 shuffles each) at the main layer.}
\label{tab:s_cxr20}
\end{table}

\section{Joint Age and Sex Matching}
\label{sec:supp_joint}
\noindent Matching on age alone left small residual costs in some settings. This section shows that pairing patients by sex as well removes all of them, so those residuals were sex shortcuts.

Matching is greedy and exact on sex, with the same 2-year age caliper, 20 matchings, and 20 shuffles as the age-only excess. We ran it for every setting except PAPILA, where age-only matching already leaves no residual and splitting 79 pairs by sex would leave too few for stable folds. Exact sex matching paired nearly every positive that age-only matching paired, so the comparison holds sample size essentially fixed.

\begin{table}[H]
\centering\footnotesize\setlength{\tabcolsep}{4pt}
\begin{tabular}{@{}llccccc@{}}
\toprule
 & & & \multicolumn{2}{c}{Age matching} & \multicolumn{2}{c}{Age and sex matching} \\
\cmidrule(lr){4-5}\cmidrule(l){6-7}
Setting & Model & Pairs & $\bar E$ (range) & $E{<}0$ & $\bar E$ (range) & $E{<}0$ \\
\midrule
ODIR-5K glaucoma & RETFound & 316 & $-.0080$ & 19/20 & $-.0003$ ($-.0073$, $+.0056$) & 10/20 \\
 & DINOv2 & 316 & $+.0010$ & 9/20 & $-.0009$ ($-.0072$, $+.0058$) & 10/20 \\
 & Dual-IFM & 316 & $-.0030$ & 13/20 & $+.0005$ ($-.0043$, $+.0072$) & 9/20 \\
CXR cardiomegaly & RAD-DINO & 1,494 & $-.0017$ & 18/20 & $+.0005$ ($-.0006$, $+.0017$) & 5/20 \\
 & DINOv2-B & 1,494 & $-.0059$ & 20/20 & $.0000$ ($-.0022$, $+.0011$) & 8/20 \\
 & DINOv2-L & 1,494 & $-.0080$ & 20/20 & $-.0002$ ($-.0017$, $+.0007$) & 12/20 \\
CXR effusion & RAD-DINO & 3,030 & $+.0003$ & 2/20 & $+.0004$ ($-.0000$, $+.0007$) & 1/20 \\
 & DINOv2-B & 3,030 & $-.0004$ & 18/20 & $-.0004$ ($-.0010$, $+.0001$) & 17/20 \\
 & DINOv2-L & 3,030 & $.0000$ & 9/20 & $+.0003$ ($-.0001$, $+.0010$) & 2/20 \\
\bottomrule
\end{tabular}
\caption{\textbf{Every residual left by age matching disappears under joint age and sex matching.} Pairs: matched pairs under joint matching (age-only matching paired 316, 1,496, and 3,033). DINOv2-B's effusion residual is consistent but below 0.001 under both matchings.}
\label{tab:s_joint}
\end{table}

\noindent\textbf{Sensitivity of the verdict rule.} Under age-only matching, tightening the overlap rule from ($|\bar E|\ge0.001$, $n_-\ge18$) to ($|\bar E|\ge0.002$, $n_-\ge18$) or ($|\bar E|\ge0.001$, $n_-\ge19$) changes one of 12 verdicts: RAD-DINO cardiomegaly ($-0.0017$, 18 of 20) becomes weak. Under joint matching no setting meets either rule, so the final verdicts do not depend on the threshold.

\section{Known-Answer Experiments: Details}
\label{sec:supp_known}
\noindent To check the test, we built cases where we know the right answer and asked whether the test finds it. This section describes how each case was built and reports how often the test raised a false alarm.

All known-answer runs use RETFound or DINOv2 features at the main layer on ODIR-5K, with 10 matchings and 10 shuffles each.

\noindent\textbf{Added confounding.} We kept all 316 glaucoma eyes and drew 2,500 control eyes without replacement, either uniformly or with weights $\exp(-\lambda a)$, with $\lambda$ chosen by bisection to make controls younger. This widened the mean age gap from 5.3 to 9.3 years. The same two subsamples were used for both models.

\noindent\textbf{Planted labels.} With $D$ the two erased directions computed on all features, we drew a random direction $u$, removed its components along $D$, and set $y=\mathbb 1[z(Xu)+\beta z(\cdot)+\varepsilon\ge q_{0.9}]$ with $\varepsilon\sim\mathcal N(0,1)$. Here $\beta{=}0$ gives no demographic signal, and $\beta{=}0.8$ applied to standardized true age gives a pure age shortcut. A first version built $u$ in standardized feature space. Because the erasure acts on raw features, that $u$ kept part of the age direction, so we report only the corrected construction in raw feature space. For DINOv2, the false excess of the planted age shortcut persisted with an exact age match ($c{=}0$: $-0.0049$, 8 of 10; $c{=}1$: $-0.0061$, 10 of 10), so it does not come from age differences left inside pairs.

\noindent\textbf{Calibration over random draws.} To measure the false-positive rate, we repeated the no-signal and pure-age-shortcut labels with five further random disease directions per model. For RETFound, no draw met the overlap rule for either label: the no-signal excess averaged $+0.0016$ (standard deviation 0.0019, minimum $-0.0003$), and the shortcut excess averaged $+0.0015$ (standard deviation 0.0009, minimum $+0.0006$). For DINOv2, no-signal labels again never met the rule (mean $+0.0008$, standard deviation 0.0012, minimum $-0.0006$), but the shortcut label met it in 2 of the 5 new draws ($-0.0033$ with 10 of 10 matchings below the null, and $-0.0015$ with 8 of 10), for 3 of 6 draws including the original (mean over the new draws $-0.0011$, standard deviation 0.0015). The planted shortcut is strong: it creates an age gap of about 10 years between positives and negatives, twice the gap for glaucoma on ODIR-5K. We therefore report a resolution of 0.005 AUC and treat smaller consistent residuals as inconclusive.

\noindent\textbf{Planted overlap.} Two constructions were tested. Defining the label from the continuous value $z(X\hat d_a)$ gave no detectable excess ($+0.0013$, 3 of 10, for RETFound). A two-direction projection removes that value, but the probe rebuilds most of it from correlated directions. Shifting the features of diseased eyes, $X\leftarrow X+\delta\,\sigma_a\,y\,\hat d_a^\top$, where $\sigma_a$ is the standard deviation of $X\hat d_a$, plants a disease signature along the age direction that does not depend on true age. This version is detected, with an excess that grows with $\delta$, on both ODIR-5K and chest X-ray features (\cref{tab:known}). Together, the two results show that the test detects overlap in the class means along the demographic directions, the quantity the erasure removes, and that it is conservative for overlap spread across correlated directions.

\section{Fine-tuning: Per-Seed Results}
\label{sec:supp_ft}
\noindent Each fine-tuning run was repeated with three random seeds. The results agree across seeds on both datasets.

\begin{table}[H]
\centering\footnotesize\setlength{\tabcolsep}{2.4pt}
\begin{tabular}{@{}lccccccc@{}}
\toprule
Run & AUC & AUC matched & $\rho$ & Score$\to$60 & Full cost [CI] & Matched cost [CI] & $p$ \\
\midrule
\multicolumn{8}{@{}l}{\emph{PAPILA} (79 pairs)} \\
seed 0 & .837 & .704 & .467 & .699 & $-.041$ [$-.074$, $-.011$] & $+.014$ [$-.002$, $.032$] & .74 \\
seed 1 & .827 & .684 & .499 & .716 & $-.040$ [$-.073$, $-.008$] & $+.008$ [$-.009$, $.027$] & .57 \\
seed 2 & .834 & .704 & .465 & .699 & $-.041$ [$-.072$, $-.010$] & $+.016$ [$.002$, $.032$] & .77 \\
\midrule
\multicolumn{8}{@{}l}{\emph{ODIR-5K} (316 pairs)} \\
seed 0 & .910 & .906 & .150 & .565 & $-.012$ [$-.019$, $-.004$] & $-.002$ [$-.011$, $.008$] & .38 \\
seed 1 & .912 & .906 & .145 & .562 & $-.015$ [$-.024$, $-.007$] & $+.001$ [$-.009$, $.010$] & .44 \\
seed 2 & .912 & .907 & .152 & .567 & $-.011$ [$-.019$, $-.004$] & $+.000$ [$-.009$, $.010$] & .46 \\
\bottomrule
\end{tabular}
\caption{\textbf{Fine-tuned RETFound, per seed.} AUC and AUC matched: end-to-end scores on all eyes and on age-matched eyes (one matching). $\rho$ and Score$\to$60: score leakage in non-glaucoma eyes. Costs: the inside test on fine-tuned features. Across seeds, fold AUCs ranged from 0.787 to 0.899 on PAPILA and from 0.875 to 0.937 on ODIR-5K.}
\label{tab:s_ft}
\end{table}

\section{Sparse-Autoencoder Diagnostics}
\label{sec:supp_sae}
\noindent This section asks whether individual SAE units respond to both a disease and age or sex. Many do, which is a property of the features and does not by itself mean that predictions rely on demographics.

\begin{figure}[H]
\centering
\includegraphics[width=0.55\linewidth]{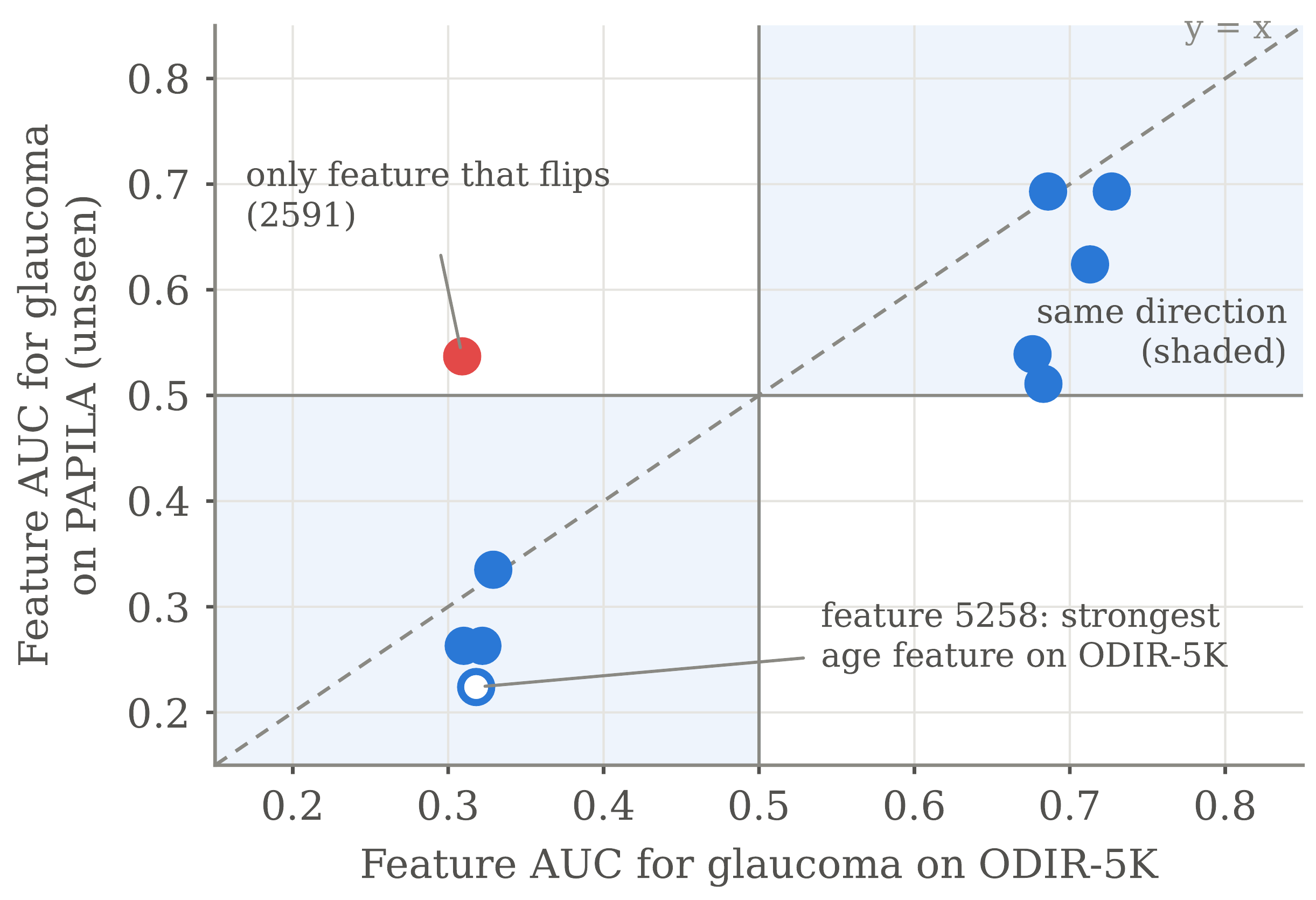}
\caption{\textbf{Transfer of RETFound's ten strongest ODIR-5K glaucoma SAE features to PAPILA}, which the autoencoder never saw. Each point is one feature's single-feature AUC for glaucoma on each dataset; points in the shaded quadrants predict glaucoma in the same direction on both. Nine of ten keep their direction. Feature 5258 (hollow), the strongest age feature on ODIR-5K, becomes one of the strongest glaucoma features on PAPILA, where age and diagnosis are confounded.}
\label{fig:s_transfer}
\end{figure}

On held-out patients the SAEs explained 0.942 (RETFound), 0.896 (DINOv2), and 0.981 (Dual-IFM) of the variance, with 27\%, 7\%, and 40\% of features never active. For each label we counted how many of a model's 20 (or 50) features most predictive of the label were also among its features most predictive of age${\ge}60$ or of sex, and compared the total with 1,000 patient-level demographic shuffles (\cref{tab:s_sae}). The shuffle also breaks real links between age and disease. This test therefore shows that single units mix the two kinds of information, but not that predictions depend on them. Applied unchanged to PAPILA, 9 of RETFound's 10 strongest ODIR-5K glaucoma features kept the sign of their association. Two of the five strongest PAPILA glaucoma features were also among the five strongest age features, and a third was the strongest age feature on ODIR-5K, which shows the PAPILA shortcut at the level of individual features. Whole probes transferred less well: a glaucoma probe trained on ODIR-5K and applied to PAPILA reached AUCs of 0.627 (RETFound), 0.684 (DINOv2), and 0.709 (Dual-IFM), and adding ODIR-5K's demographic projection lowered these to 0.595, 0.663, and 0.706 (\cref{fig:s_transfer}). Used for debiasing, dropping SAE features correlated with sex or age (threshold 0.05) barely changed how well sex could be read (RETFound 0.559 to 0.551) and cost up to 0.06 disease AUC (RETFound AMD 0.823 to 0.762), a worse trade-off than linear projection.

\begin{table}[H]
\centering\footnotesize\setlength{\tabcolsep}{2.5pt}
\adjustbox{max width=\textwidth}{%
\begin{tabular}{@{}lcccccc@{}}
\toprule
 & \multicolumn{3}{c}{Age ${\ge}60$: observed (chance), $p$} & \multicolumn{3}{c}{Sex: observed (chance), $p$} \\
\cmidrule(lr){2-4}\cmidrule(l){5-7}
Model & Top 20 & Top 50 & Glauc.\ top 20 & Top 20 & Top 50 & Glauc.\ top 20 \\
\midrule
RETFound & 14 (1.5), .001 & 59 (8.8), .001 & 7 (0.3) & 11 (1.4), .003 & 32 (8.4), .006 & 2 (0.3) \\
DINOv2 & 5 (0.8), .013 & 35 (4.7), .001 & 2 (0.1) & 5 (0.8), .019 & 27 (4.5), .001 & 2 (0.1) \\
Dual-IFM & 20 (1.0), .001 & 61 (5.3), .001 & 6 (0.2) & 9 (0.8), .004 & 34 (5.0), .004 & 1 (0.2) \\
\bottomrule
\end{tabular}}
\caption{\textbf{SAE feature overlap, top 20 and top 50.} Number of each model's strongest disease features (summed over seven ODIR-5K labels) that are also among its strongest age or sex features; chance level from 1,000 patient-level shuffles in parentheses, followed by $p$.}
\label{tab:s_sae}
\end{table}

\section{Outside-in Audit}
\label{sec:supp_out}
\noindent This audit uses only the model's output scores, as a hospital without access to the features could. Age adds far more to glaucoma prediction on PAPILA than on ODIR-5K.

\begin{table}[H]
\centering\footnotesize\setlength{\tabcolsep}{3pt}
\begin{tabular}{@{}llcccc@{}}
\toprule
Dataset & Model & Gain over age+sex [95\% CI] & OR score & OR age \\
\midrule
ODIR-5K & RETFound & .196 [.150, .243] & 3.68 & 1.33 \\
 & DINOv2 & .215 [.174, .256] & 3.20 & 1.40 \\
 & Dual-IFM & .195 [.152, .239] & 3.13 & 1.43 \\
PAPILA & RETFound & .101 [.045, .162] & 3.49 & 2.44 \\
 & DINOv2 & .096 [.039, .160] & 3.69 & 2.20 \\
 & Dual-IFM & .080 [.027, .133] & 3.21 & 2.34 \\
\bottomrule
\end{tabular}
\caption{\textbf{Glaucoma, score only.} Gain: AUC added by the score over an age-and-sex baseline. Odds ratios are per standard deviation, from a logistic regression on age, sex, and the logit of the score, with standard errors clustered by patient.}
\label{tab:s_out}
\end{table}
\noindent On ODIR-5K, AUC was similar across all six age-by-sex strata for every label. On PAPILA, glaucoma prevalence rose from 6 to 7\% below age 60 to 42 to 47\% at 70 and above, and RETFound's AUC fell from 0.929 (women under 60) to between 0.70 and 0.73 at 70 and above, with wide intervals. The gap in disease AUC between men and women stayed at or below 0.04, and erasure did not change it.

\section{Clinician Review}
\label{sec:supp_clin}
\noindent An ophthalmologist examined what the model's units respond to and which healthy eyes the model scores as most glaucomatous, without knowing the labels.

A practicing ophthalmologist reviewed two blinded packets after giving written informed consent, under a protocol approved by an institutional review board.

\noindent\textbf{Feature packet.} From the RETFound SAE on ODIR-5K we selected, for each of glaucoma, cataract, myopia, and AMD, the four features whose single-feature AUC was farthest from 0.5, plus three features for age${\ge}60$. The 19 features appeared in random order under neutral codes. Each page showed the six top-activating images, with boxes where activation reached at least half of its maximum. For each feature the reviewer described the shared structure, chose a location, rated her confidence, and judged whether the feature was clinically meaningful. She did not see our automated labels or the reason each feature was selected. She judged 13 of the 19 meaningful and located them on the optic disc, the vessels and arcades, the macula, areas of atrophy, exudates, and the periphery. Our automated labeling had independently flagged four of the six rejected features as image-border or image-quality features.

\noindent\textbf{Eye packet.} Among healthy PAPILA eyes, taking one eye per patient, we selected the 10 with the highest and the 10 with the lowest out-of-fold RETFound glaucoma scores. Age, sex, diagnosis, and scores were hidden from the reviewer.

\begin{table}[H]
\centering\footnotesize
\begin{tabular}{@{}lccc@{}}
\toprule
 & High score & Low score & Test ($p$) \\
\midrule
Judged suspicious & 4 of 10 & 1 of 10 & Fisher, one-sided (.15) \\
Mean age (years) & 68.1 & 52.9 & Mann--Whitney (.03) \\
Mean cup-to-disc (estimate) & 0.35 & 0.37 & Mann--Whitney (.87) \\
Fair or poor image quality & 5 of 8 & 0 of 9 & Fisher, two-sided (.009) \\
\bottomrule
\end{tabular}
\caption{\textbf{Blinded review of 20 healthy PAPILA eyes.} High-score eyes are older and of poorer image quality, and their cups are no larger.}
\label{tab:s_clin}
\end{table}

\section{Multiple Comparisons}
\label{sec:supp_mc}
\noindent Running many tests raises the chance of a false positive. This section checks that the main results survive a standard correction.

Applying the Benjamini--Hochberg procedure at 5\% to the 24 retinal projection-cost tests (ODIR-5K labels and PAPILA), all 15 results with $p{=}0.0099$ remain significant, including every PAPILA result. Only DINOv2's ODIR-5K costs for glaucoma ($p{=}0.040$) and myopia ($p{=}0.050$) lose significance. With 100 nulls the smallest attainable $p$ is 0.0099, so the strongest results cannot be ranked against one another.

%% file: main.bbl
\begin{thebibliography}{22}
\providecommand{\natexlab}[1]{#1}
\providecommand{\url}[1]{\texttt{#1}}
\expandafter\ifx\csname urlstyle\endcsname\relax
  \providecommand{\doi}[1]{doi: #1}\else
  \providecommand{\doi}{doi: \begingroup \urlstyle{rm}\Url}\fi

\bibitem[Alain and Bengio(2016)]{alain2016probes}
Guillaume Alain and Yoshua Bengio.
\newblock Understanding intermediate layers using linear classifier probes.
\newblock \emph{arXiv preprint arXiv:1610.01644}, 2016.

\bibitem[Belrose et~al.(2023)Belrose, Schneider-Joseph, Ravfogel, Cotterell,
  Raff, and Biderman]{belrose2023leace}
Nora Belrose, David Schneider-Joseph, Shauli Ravfogel, Ryan Cotterell, Edward
  Raff, and Stella Biderman.
\newblock {LEACE}: Perfect linear concept erasure in closed form.
\newblock In \emph{Advances in Neural Information Processing Systems
  (NeurIPS)}, 2023.

\bibitem[Brown et~al.(2023)Brown, Tomasev, Freyberg, Liu, Karthikesalingam, and
  Schrouff]{brown2023short}
Alexander Brown, Nenad Tomasev, Jan Freyberg, Yuan Liu, Alan Karthikesalingam,
  and Jessica Schrouff.
\newblock Detecting shortcut learning for fair medical {AI} using shortcut
  testing.
\newblock \emph{Nature Communications}, 14:\penalty0 4314, 2023.

\bibitem[Engelmann and Bernabeu(2025)]{engelmann2025retfoundgreen}
Justin Engelmann and Miguel~O. Bernabeu.
\newblock Training a high-performance retinal foundation model with
  half-the-data and 400 times less compute.
\newblock \emph{Nature Communications}, 16, 2025.

\bibitem[Gao et~al.(2024)Gao, Dupr{\'e}~la Tour, Tillman, et~al.]{gao2024sae}
Leo Gao, Tom Dupr{\'e}~la Tour, Henk Tillman, et~al.
\newblock Scaling and evaluating sparse autoencoders.
\newblock \emph{arXiv preprint arXiv:2406.04093}, 2024.

\bibitem[Geirhos et~al.(2020)Geirhos, Jacobsen, Michaelis, Zemel, Brendel,
  Bethge, and Wichmann]{geirhos2020shortcut}
Robert Geirhos, J{\"o}rn-Henrik Jacobsen, Claudio Michaelis, Richard Zemel,
  Wieland Brendel, Matthias Bethge, and Felix~A. Wichmann.
\newblock Shortcut learning in deep neural networks.
\newblock \emph{Nature Machine Intelligence}, 2\penalty0 (11):\penalty0
  665--673, 2020.

\bibitem[Gichoya et~al.(2022)Gichoya, Banerjee, Bhimireddy, Burns, Celi, Chen,
  Correa, Dullerud, Ghassemi, Huang, et~al.]{gichoya2022race}
Judy~Wawira Gichoya, Imon Banerjee, Ananth~Reddy Bhimireddy, John~L. Burns,
  Leo~Anthony Celi, Li-Ching Chen, Ramon Correa, Natalie Dullerud, Marzyeh
  Ghassemi, Shih-Cheng Huang, et~al.
\newblock {AI} recognition of patient race in medical imaging: a modelling
  study.
\newblock \emph{The Lancet Digital Health}, 4\penalty0 (6):\penalty0
  e406--e414, 2022.

\bibitem[Glocker et~al.(2023)Glocker, Jones, Roschewitz, and
  Winzeck]{glocker2023cxrfm}
Ben Glocker, Charles Jones, M{\'e}lanie Roschewitz, and Stefan Winzeck.
\newblock Risk of bias in chest radiography deep learning foundation models.
\newblock \emph{Radiology: Artificial Intelligence}, 5\penalty0 (6):\penalty0
  e230060, 2023.

\bibitem[Kovalyk et~al.(2022)Kovalyk, Morales-S{\'a}nchez, Verd{\'u}-Monedero,
  Sell{\'e}s-Navarro, Palaz{\'o}n-Cabanes, and
  Sancho-G{\'o}mez]{kovalyk2022papila}
Oleksandr Kovalyk, Juan Morales-S{\'a}nchez, Rafael Verd{\'u}-Monedero,
  Inmaculada Sell{\'e}s-Navarro, Ana Palaz{\'o}n-Cabanes, and Jos{\'e}-Luis
  Sancho-G{\'o}mez.
\newblock {PAPILA}: Dataset with fundus images and clinical data of both eyes
  of the same patient for glaucoma assessment.
\newblock \emph{Scientific Data}, 9:\penalty0 291, 2022.

\bibitem[Le et~al.(2024)Le, Shen, Patel, Shah, Sanghavi, Martin, Eng, Shenker,
  Padigela, Biju, et~al.]{le2024pathsae}
Nhat~Minh Le, Ciyue Shen, Neel Patel, Chintan Shah, Darpan Sanghavi, Blake
  Martin, Alfred Eng, Daniel Shenker, Harshith Padigela, Raymond Biju, et~al.
\newblock Learning biologically relevant features in a pathology foundation
  model using sparse autoencoders.
\newblock \emph{arXiv preprint arXiv:2407.10785}, 2024.

\bibitem[Mensah et~al.(2026)Mensah, Roa, Djoumessi, and Berens]{dualifm2026}
Samuel~Ofosu Mensah, Camila Roa, Kerol Djoumessi, and Philipp Berens.
\newblock Towards interpretable foundation models for retinal fundus images.
\newblock \emph{arXiv preprint arXiv:2603.18846}, 2026.

\bibitem[Munk et~al.(2021)Munk, Kurmann, M{\'a}rquez-Neila, Zinkernagel, Wolf,
  and Sznitman]{munk2021patient}
Marion~R. Munk, Thomas Kurmann, Pablo M{\'a}rquez-Neila, Martin~S. Zinkernagel,
  Sebastian Wolf, and Raphael Sznitman.
\newblock Assessment of patient specific information in the wild on fundus
  photography and optical coherence tomography.
\newblock \emph{Scientific Reports}, 11:\penalty0 8621, 2021.

\bibitem[Nielsen et~al.(2025)Nielsen, Stanley, Wilms, and
  Forkert]{nielsen2025retinalage}
Christopher Nielsen, Emma A.~M. Stanley, Matthias Wilms, and Nils~D. Forkert.
\newblock Assessment of demographic bias in retinal age prediction machine
  learning models.
\newblock \emph{Frontiers in Artificial Intelligence}, 8:\penalty0 1653153,
  2025.

\bibitem[Oquab et~al.(2024)Oquab, Darcet, Moutakanni, et~al.]{oquab2024dinov2}
Maxime Oquab, Timoth{\'e}e Darcet, Th{\'e}o Moutakanni, et~al.
\newblock {DINOv2}: Learning robust visual features without supervision.
\newblock \emph{Transactions on Machine Learning Research}, 2024.

\bibitem[{Peking University}(2019)]{odir2019}
{Peking University}.
\newblock Ocular disease intelligent recognition ({ODIR-2019}) dataset.
\newblock International Competition on Ocular Disease Intelligent Recognition,
  2019.

\bibitem[P{\'e}rez-Garc{\'i}a et~al.(2025)P{\'e}rez-Garc{\'i}a, Sharma,
  Bond-Taylor, Bouzid, Salvatelli, Ilse, Bannur, Castro, Schwaighofer, Lungren,
  et~al.]{perezgarcia2025raddino}
Fernando P{\'e}rez-Garc{\'i}a, Harshita Sharma, Sam Bond-Taylor, Kenza Bouzid,
  Valentina Salvatelli, Maximilian Ilse, Shruthi Bannur, Daniel~C. Castro,
  Anton Schwaighofer, Matthew~P. Lungren, et~al.
\newblock Exploring scalable medical image encoders beyond text supervision.
\newblock \emph{Nature Machine Intelligence}, 7:\penalty0 119--130, 2025.

\bibitem[Ravfogel et~al.(2020)Ravfogel, Elazar, Gonen, Twiton, and
  Goldberg]{ravfogel2020inlp}
Shauli Ravfogel, Yanai Elazar, Hila Gonen, Michael Twiton, and Yoav Goldberg.
\newblock Null it out: Guarding protected attributes by iterative nullspace
  projection.
\newblock In \emph{Proceedings of the 58th Annual Meeting of the Association
  for Computational Linguistics (ACL)}, pages 7237--7256, 2020.

\bibitem[Ravfogel et~al.(2022)Ravfogel, Vargas, Goldberg, and
  Cotterell]{ravfogel2022kernel}
Shauli Ravfogel, Francisco Vargas, Yoav Goldberg, and Ryan Cotterell.
\newblock Adversarial concept erasure in kernel space.
\newblock In \emph{Proceedings of the 2022 Conference on Empirical Methods in
  Natural Language Processing (EMNLP)}, pages 6034--6055, 2022.

\bibitem[Smith et~al.(2025)Smith, Rajamanoharan, Conmy, McDougall, Kram{\'a}r,
  Lieberum, Shah, and Nanda]{smith2025saeneg}
Lewis Smith, Senthooran Rajamanoharan, Arthur Conmy, Callum McDougall,
  J{\'a}nos Kram{\'a}r, Tom Lieberum, Rohin Shah, and Neel Nanda.
\newblock Negative results for {SAEs} on downstream tasks and deprioritising
  {SAE} research ({GDM} mech interp team progress update \#2).
\newblock AI Alignment Forum, 2025.

\bibitem[Tham et~al.(2014)Tham, Li, Wong, Quigley, Aung, and
  Cheng]{tham2014glaucoma}
Yih-Chung Tham, Xiang Li, Tien~Y. Wong, Harry~A. Quigley, Tin Aung, and
  Ching-Yu Cheng.
\newblock Global prevalence of glaucoma and projections of glaucoma burden
  through 2040: a systematic review and meta-analysis.
\newblock \emph{Ophthalmology}, 121\penalty0 (11):\penalty0 2081--2090, 2014.

\bibitem[Wang et~al.(2017)Wang, Peng, Lu, Lu, Bagheri, and
  Summers]{wang2017chestxray8}
Xiaosong Wang, Yifan Peng, Le Lu, Zhiyong Lu, Mohammadhadi Bagheri, and
  Ronald~M. Summers.
\newblock {ChestX-ray8}: Hospital-scale chest {X}-ray database and benchmarks
  on weakly-supervised classification and localization of common thorax
  diseases.
\newblock In \emph{Proceedings of the IEEE Conference on Computer Vision and
  Pattern Recognition (CVPR)}, pages 2097--2106, 2017.

\bibitem[Zhou et~al.(2023)Zhou, Chia, Wagner, Ayhan, Williamson, Struyven, Liu,
  Xu, Lozano, Woodward-Court, et~al.]{zhou2023retfound}
Yukun Zhou, Mark~A. Chia, Siegfried~K. Wagner, Murat~S. Ayhan, Dominic~J.
  Williamson, Robbert~R. Struyven, Timing Liu, Moucheng Xu, Mateo~G. Lozano,
  Peter Woodward-Court, et~al.
\newblock A foundation model for generalizable disease detection from retinal
  images.
\newblock \emph{Nature}, 622\penalty0 (7981):\penalty0 156--163, 2023.

\end{thebibliography}
